\documentclass[sn-apa]{sn-jnl}

\usepackage{float}
\usepackage{multirow}%
\usepackage{amsmath,amssymb,amsfonts}%
\usepackage{amsthm}%
\usepackage{mathrsfs}%
\usepackage[title]{appendix}%
\usepackage{textcomp}%
\usepackage{manyfoot}%
\usepackage{booktabs}%
\usepackage{algorithm}%
\usepackage{algorithmicx}%
\usepackage{algpseudocode}%
\usepackage{listings}%
\usepackage{subcaption}
\usepackage{multicol}
\usepackage{color}

\DeclareCaptionLabelSeparator{none}{ }
\usepackage{tabularx}
\usepackage[symbol*]{footmisc}

\usepackage{fancyhdr}

\theoremstyle{thmstyleone}%
\theoremstyle{thmstyletwo}%

\theoremstyle{thmstylethree}%

\usepackage{graphicx}
\usepackage[dvipsnames]{xcolor}  
\definecolor{myviolet}{rgb}{255, 0, 255}

\begin{document}

\title[Article Title]{Inductive Biases for Zero-shot Systematic Generalization in Language-informed Reinforcement Learning}


\author[1]{\fnm{Negin} \sur{Hashemi Dijujin}}\email{n.hashemi94@sharif.edu}

\author[2]{\fnm{Seyed Roozbeh} \sur{Razavi Rohani}}\email{razavii.roozbeh@gmail.com}

\author[1]{\fnm{Mahdi} \sur{Samiee}}\email{mm.samiei@sharif.edu}

\author*[1]{\fnm{Mahdieh} \sur{Soleymani Baghshah}}\email{soleymani@sharif.edu}

\affil*[1]{\orgdiv{Computer Engineering Department}, \orgname{Sharif University of Technology}, \orgaddress{\street{Azadi}, \city{Tehran}, \postcode{1458889694}, \state{Tehran}, \country{Iran}}}

\affil*[2]{\orgdiv{Alumni of Computer Engineering Department}, \orgname{Sharif University of Technology}}

\abstract{Sample efficiency and systematic generalization are two long-standing challenges in reinforcement learning. Previous studies have shown that involving natural language along with other observation modalities can improve generalization and sample efficiency due to its compositional and open-ended nature. However, to transfer these properties of language to the decision-making process, it is necessary to establish a proper language grounding mechanism. One approach to this problem is applying inductive biases to extract fine-grained and informative representations from the observations, which makes them more connectable to the language units. We provide architecture-level inductive biases for modularity and sparsity mainly based on \textit{Neural Production Systems (NPS)}. Alongside NPS, we assign a central role to memory in our architecture. It can be seen as a high-level information aggregator which feeds policy/value heads with comprehensive information and simultaneously guides selective attention in NPS through attentional feedback. Our results in the BabyAI environment suggest that the proposed model's systematic generalization and sample efficiency are improved significantly compared to previous models. An extensive ablation study on variants of the proposed method is conducted, and the effectiveness of each employed technique on generalization, sample efficiency, and training stability is specified.}




\maketitle

\section{Introduction}\label{sec1}

Language as a unique communication and thinking system allows the recombining abstract units to create new meanings in countless ways according to specific rules. This property, called \textit{compositional generalization} or \textit{systematic generalization}, underlies many of our cognitive abilities, including our ability to reason, plan, and imagine \citep{ito2022compositional, chomsky2014aspects, lake_generalization_2018, lake_human_2019, berko1958child, mitchell_composition_2010}, and can improve generalization properties of deep architectures once 
incorporated effectively in models \citep{ito2022compositional}. Many investigations have been conducted based on this hypothesis to transfer the knowledge and structure of language to the deep models \citep{lake_human_2019, chen_compositional_2020, keysers_measuring_2020, akyurek_learning_2021, lake_compositional_2019}.

In reinforcement learning settings, language-informed studies \citep{luketina_survey_2019, roder_embodied_2021, geffner_target_2022} aim to assist agents by incorporating natural language sentences as an additional input besides visual observation. By leveraging language, such agents can learn complex tasks more sample efficiently and generalize to unseen tasks more effectively \citep{luketina_survey_2019, cao2020babyai++, chevalier-boisvert_babyai_2018, goyal_using_2019}. This is particularly useful in settings where the tasks are too complex to be defined by simple reward functions \citep{goyal_using_2019, fu_language_2018, mirchandani2021ella} or where human guidance is necessary for the agent to perform well \citep{wang_grounding_2021, zhong_rtfm_2020, co-reyes_guiding_2018, chen_ask_2021, hill_human_2020}. It is known that effective learning in language-informed reinforcement learning depends on the agent's ability to ground linguistic concepts in the observation \citep{cao2020babyai++, wang_grounding_2021, akakzia_grounding_2021, colas_language_2020}. While the compositional nature of the input language enhances generalization \citep{goyal_using_2019, fu_language_2018, misra_mapping_2017}, it is not enough by itself to solve the benchmarked tasks \citep{cao2020babyai++, chevalier-boisvert_babyai_2018, kuttler_nethack_2020}.

Although some recent studies have shown additional inductive biases such as modularity and sparse processing of information can help to boost the capacity for compositional generalization \citep{bahdanau_systematic_2019, spilsbury_compositional_2022, hein_minimal_2022}, these ideas have not already been employed in RL problems. Yet, language-informed RL studies only leverage techniques such as cross-attention \citep{cao2020babyai++, wang_grounding_2021}, modulation \citep{zhong_rtfm_2020, perez_film_2018} or concatenation \citep{chevalier-boisvert_babyai_2018} to fuse language with other raw inputs. In this study, we highlight the role of modularity and sparse interactions for compositional generalization in language-informed RL. By utilizing proper structural inductive biases into the encoder part of the policy/value function, we provide a modular network that factorizes knowledge about interacting objects or entities in the form of differentiable condition-action rules. More specifically, we employ Neural Production Systems (NPS) \citep{alias_parth_goyal_neural_2021}, consisting of a set of encoded rules that can be applied to specific input parts, called slots, in a sparse manner since the direct slot-to-slot interactions are not further required. We also enrich NPS with two techniques to take better advantage of the language modality alongside the inputs processed by NPS: \textit{language entrance}, and \textit{memory feedback}. In doing so, we transfer the desired properties of language to the model to promote its modularity and sparsity, which may lead to compositionality in the network representations that are useful for generalization. 

According to neuroscience studies, \textit{Prefrontal Cortex (PFC)} is involved in \textit{Working Memory (WM)}, which describes having the ability to keep and manipulate information that is no longer accessible in the environment \citep{wang2018prefrontal, rohani_bimrl_2022}. It is also involved in natural language understanding \citep{gonzalez-garcia_frontoparietal_2021, muhle-karbe_neural_2017}, and in \textit{selective attention} which refers to the functions that prioritize and select information to guide adaptive behavior \citep{nobre2019premembering, radulescu_holistic_2019, paneri_top-down_2017}. As we will see in Section \ref{architecture}, by involving language information through memory feedback, we developed a process, like the one that happens in selective visual attention between PFC and mid-level visual processing regions, where high-level information in PFC is employed to attend to specific parts of visual input through attentional feedback \citep{radulescu_holistic_2019}.

We run our experiments on several levels in the BabyAI environment \citep{chevalier-boisvert_babyai_2018}, a rich and light-weighted testbed for instruction-following decision-making agents which imposes challenges like complex goals, sparse rewards, and multi-task settings in various difficulty levels. Our results on a systematic training/testing split indicate a significantly superior performance of the proposed method compared to previous encoders in the literature. According to our ablation study, the proposed additional techniques outperform the strong base models with improved training stability, total return, generalization gap, and sample efficiency. 
The summary of our contributions is that:
\begin{itemize}
 \item We emphasize the importance
 of modularity and sparsity in RL settings for systematic generalization.
 \item We propose a modular architecture based on NPS for the observation encoding which provides a better framework for incorporation of the language instruction. 
 \item We introduce a memory feedback which utilizes an aggregation of observations encoding and the language instruction in the attention-based context or rule selection process. 
 We also state the neuroscientific studies supporting the proposed memory feedback mechanism.
 \item Experimental results showcase the capability of the proposed model for computational generalization compared to the previous studies.
\end{itemize}

\section{Background}
\label{background}
We build off the NPS \citep{alias_parth_goyal_neural_2021}, which is a neural versions of Production Systems \citep{lovett_thinking_2005}, introduced in the late 1960s as a standard tool for describing how human beings think. A production system consists of some modular and abstract rules. Each rule is a pair of condition-action mechanisms, and its action applies to the input only when the corresponding condition is met. This framework provides sufficient conditions for representing knowledge through production rules. Recently, \citep{alias_parth_goyal_neural_2021} have modeled these rules in a neural way. More precisely, actions are specified with neural networks, mainly MLPs, and conditions are represented by vectors of trainable parameters. Thus, NPS is an end-to-end differentiable neural network involving inductive biases of production systems.

Now, we describe the architecture of the NPS since it lies at the heart of our study. 
The NPS includes $N$ modular rules ${R_1,...,R_N}$ where $R_i = (\hat{R}_i, MLP_i)$ and maps the input $x_t$ to a set of entities or slots ${V_1^t,...,V_M^t}$. Then for a specific slot, called the primary slot ($V^t_p$), a rule is selected to be applied on through a competitive bottleneck resulting from the attention mechanism. More precisely, to select a rule for the primary slot $V^t_p$, we consider
\begin{align}
q_p&=V^t_p W^q_r\\ 
k_i&=R_i W^k_r\text{\hspace{1em}}(i=1,...,N)\\
r &= \arg\max_i (q_p^T k_i +\gamma) \text{\hspace{1em}} \gamma \sim Gumbel(0,1) \label{rule-selection} 
\end{align}
where the $q_p$ is the query, \textcolor{black}{$W^q_r$ and $W^k_r$ are projection matrices,} and the $k_i$s are keys of attention in Eq. \ref{rule-selection} which is a noisy rule matching \citep{alias_parth_goyal_neural_2021}. Moreover, to apply the selected rule $r$ on the slot $V^t_p$, in addition to $V^t_p$, the related context as a contextual slot $V^t_c$ which is specified using another attention mechanism, is also fed to $MLP_r$. In fact, this contextual slot is found through the attention formulated as
\begin{align}
 q_p&=V^t_p W^q_c\\
 k_j&=V^t_j W^k_c (j=1, ..., M)\\
 c&= \arg\max_j (q_p^T k_j +\gamma) \text{\hspace{1em}} \label{context-selection} \gamma \sim Gumbel(0,1) 
\end{align}
\textcolor{black}{where $W^q_c$ and $W^k_c$ are projection matrices for context selection attention,} according to \citep{alias_parth_goyal_neural_2021}. The primary slot concatenated with the contextual slot passes through the $MLP_r$ as below
\begin{align}
 out_p &= MLP_r(V^t_p \oplus V^t_c)
\end{align}
where the $out_p$ can be used to modify the state of the primary slot or passed down through the network.


The process of applying rules might be parallel or sequential.
In the parallel case, for each slot, one rule is selected and applied simultaneously at the current time step, while in the sequential case, we select only one primary slot from the whole observation at a time. The choice between these two methods depends on the input dynamics and the extent of interaction between the present entities. The parallel approach is more appropriate for the input with dense relations between the entities and vice versa. NPS has been applied to various tasks, such as performing spatial transformations on inputs or learning action-conditioned world models, but its performance on RL problems has remained underexplored.

\section{Proposed Model}
\subsection{Problem Formulation}
\label{formulation}
In this study, we are interested in multi-task instruction following sequential decision-making settings in which a natural language instruction describes the agent’s goal in a partially observable environment. Formally, we are trying to solve an augmented POMDP defined by the tuple $(S, A, O, \Omega, T, \tilde{R}, G, \tilde{\gamma})$ in which $S$ is the state space, $A$ is the action space, $O$ is the observation space, 
$\Omega: S \rightarrow O$ 
is an observation mapping function, 
$T: S\times A \rightarrow S$ 
is the state transition function, $\tilde{R}$ is the reward function for reinforcement learning setup, and $\tilde{\gamma}$ is the discount factor. Alongside these usual components in the POMDP definition, $G$ also contains all possible instructions for the environment in the augmented POMDP. 

We consider a multi-task setting where each task is recognized by a pair of initial state, $s_0$, and goal instruction, $g$. All MDP components are shared across tasks except $\tilde{R}$, which is affected by the task itself: $\tilde{R}: S \times A \times S \times G \rightarrow \mathbb{R}$. Finally, we attempt to learn a return-maximizing policy $\pi(a_t|o_t, g)$ which is conditioned on the instruction. In our experiments, we define a compositional split \textcolor{black}{on $G$ to divide it into two \textit{disjoint} sets,} $G_{train}$ and $G_{test}$, to assess the systematic generalizability of the proposed techniques. During training, the agent only sees instructions from $S \times G_{train}$ whereas tests are performed on tasks only inside $S \times G_{test}$. \textcolor{black}{So, $G_{test}$ contains tasks which remain \textit{unseen} during training to assess the zero-shot performance of the agent.} Because of the compositional nature of the language, \textcolor{black}{we expect that the model more effectively generalizes to unseen tasks by using prior knowledge included within the instructions.} 

\subsection{Architecture}
\label{architecture}
This study explores architecture-level inductive biases for compositional generalization in reinforcement learning. We choose NPS \citep{alias_parth_goyal_neural_2021} -described in Section \ref{background}- as the base model for our inductive biases. The modularity and sparsity of interactions between entities manifested by context selection for each primary slot are well-suited for our purpose of grounding natural language instructions in the agent’s representation of the world. 

\begin{figure}
 \centering
 \includegraphics[width=\textwidth]{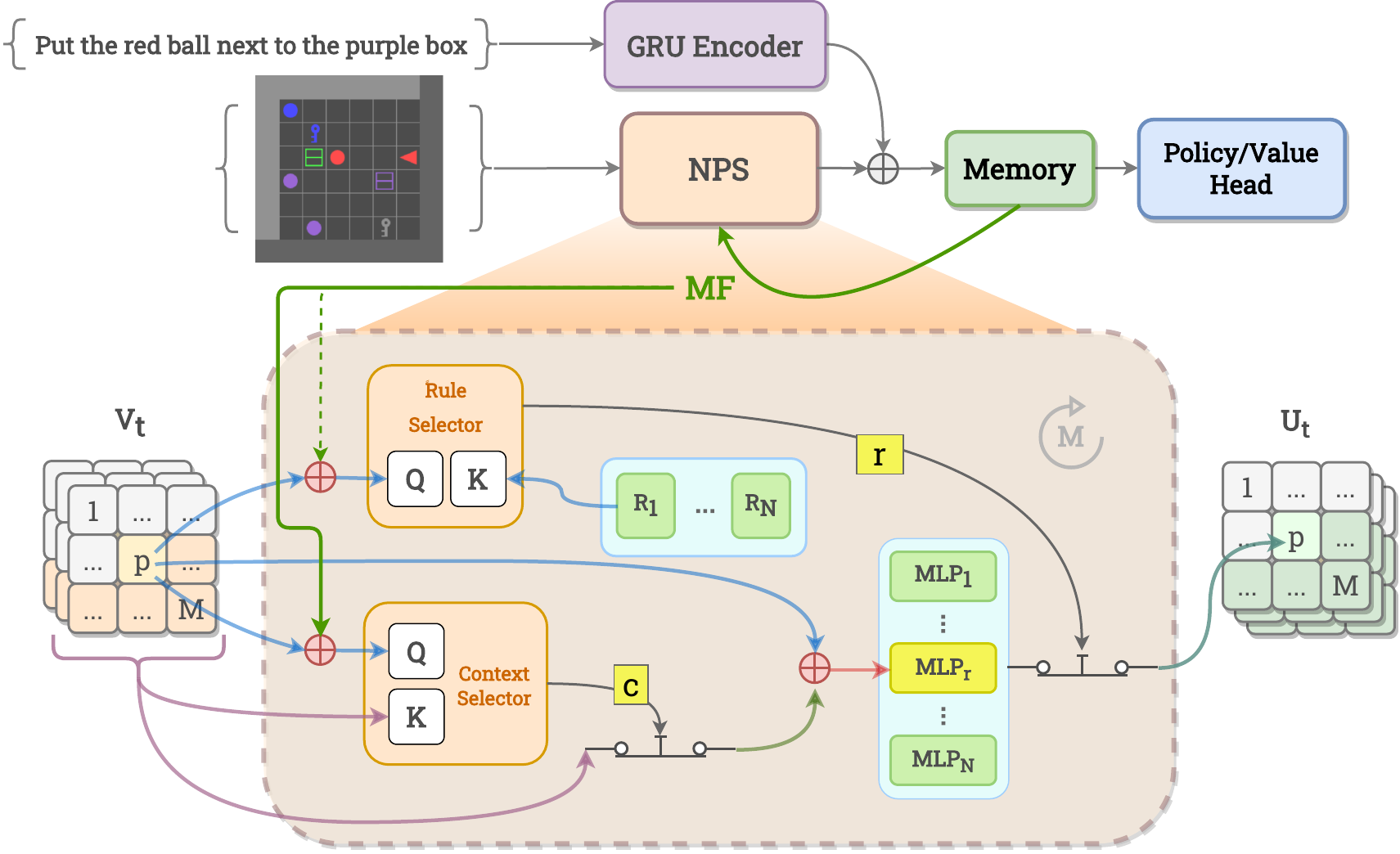}
 \caption{Overall architecture of ICMO (The switch icon (\includegraphics[width=3em]{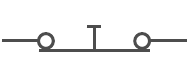}) performs index selection; for example, the output of rule selector module is an index $r$ to choose the most relevant rule action, $MLP_r$, and the left port of this switch receives the array and the right port outputs the selected item)}
 \label{ICMO}
\end{figure} 

In the rest of this section, we describe the overall architecture of the model based on NPS described in Section \ref{background} in which for an observation $o_t$ consisting of slots $V_t = \{V_1^t, ..., V_M^t\}$, we input these slots to the model. According to Fig. \ref{ICMO}, the output of the NPS for $V_t$, i.e., $U_t$, passes through a recurrent neural network called memory to obtain $h_t$ from the previous memory state $h_{t-1}$ and the encoding of the observation $U_t$. Thereafter, a policy/value head outputs actions/values given the memory’s hidden state as the input. We modify this procedure by adding two inductive biases related to \textit{memory feedback} and \textcolor{black}{\textit{language entrance}} described below.

\begin{itemize}
 \item 
\textbf{Memory Feedback:} To enrich the NPS architecture, memory feedback is incorporated into the selection mechanisms. By default, this query is the primary slot, but we also extract another representation from memory through a linear layer and concatenate it to the encoding of primary slots. More specifically, we connect the memory’s hidden state from the previous timestep ($h_{t-1}$) back to the NPS by modifying the query as

\begin{align}
 MF &:= W_m^T h_{t-1} + b_m \\
 q_p &= V_p^t \oplus MF
\end{align}

where $q_p$ replaces the query in Eq. \ref{rule-selection} or Eq. \ref{context-selection}, \textcolor{black}{and $W_m$ and $b_m$ are learnable weights}. So the query of the attention for the selection of rule or contextual slot is modified to contain the past information from agent's memory.

The intuition behind memory feedback is that the entities in the instruction may happen sparsely through the episode due to the partial observability of the environment. Memory feedback helps the agent to incorporate past information in its selective mechanisms. Remembering the previous experiences in the episode helps avoid repetitive unnecessary interactions with the environment, and reduces the episode's length. 
As it will be demonstrated in Section \ref{experiments}, this feedback connection specifically improves performance when it contains instruction information as well.

\item 
\textbf{Language Entrance:} The time-invariant task information, i.e. instruction in our setting, is considered as a condition in the architecture. There are several places in the architecture where we can enter the language as a condition: 1) the embedding of the language instruction can be considered as an input of the memory module that aggregates observations, 2) it can be fed lately to the policy/value head of the model, or 3) it can be done via an early fusion of language information with the observation at each time step. 

\textcolor{black}{Although first and second designs both can cause similar feed-forward effects by combining language with high-level representations, the first design provides richer outputs in terms of feedback by grounding the language in the memory.} Experimental results of the next section confirm that the first design is the best one combined with memory feedback. One can reason that the language instruction enters during the aggregation of the observations to process the encoded observations (prepared by NPS) with a guidance which can highlight more informative elements of the memory, i.e., specify the completed sets of sub-goals or the essential features of the current state. 

\end{itemize}

We call the proposed method \textit{\textbf{I}nstruction \textbf{C}onditioned \textbf{MO}dular network}, or \textbf{ICMO} for short and showcase its superior performance in our experiments. The proposed architecture is agnostic to the training algorithm and can work with any reinforcement learning or even imitation learning algorithms. 
It is worth noting that, due to the sparsity and high abstraction of language-informed RL tasks, the instruction is frequently not directly coupled to the observations in each time step; since a fixed instruction is considered for a whole sequence of observations as opposed to supervised vision-language tasks in which each image is paired with a text. Therefore, putting the language instruction where its level of detail is more appropriate is helpful. In this case, by extending the query to be memory-aware, the language instruction may indirectly affect the selection of rules.

From the neurocognitive point of view, the inductive biases injected in the proposed method are consistent with findings about the significant role of WM in action-oriented tasks and modularity in structural and functional aspects of the brain \citep{meunier_modular_2010, sporns_modular_2016, perich_rethinking_2020, power_functional_2011, wang_brain_2016, yang2019task}. Through the functionality lens, a highly modular, sparsely activated architecture for observation encoding, could be considered as mid-level visual processing region, which is also modulated by attentional feedback from PFC \citep{radulescu_holistic_2019} -resembled by the memory in our model- to selectively attend specific parts of input. In the proposed method, through the memory module and its role in feature selection, WM is responsible for the control information pathways that let previously learned modules dynamically combine \citep{riveland2022generalization} to fuse language and observation. In the end, aggregated information is fed to the actor-critic network, whose functionalities are associated with the striatum \citep{sutton_reinforcement_2018}, a subcortical region in the brain. For more details and connections to theories from the neuroscience side, please see Section \ref{apprelatedworks}.

\section{Experiments}
\label{experiments}
In this section, we explain our experimental setup (Section \ref{setup}) and results (Section \ref{results}). \textcolor{black}{Further analyses are stated in Discussions, Section \ref{sec:disc}.}
\subsection{Setup}
\label{setup}
Our problem setup consists of the benchmark for systematic generalization defined on BabyAI \citep{chevalier-boisvert_babyai_2018} environment with an additional train/test split (Section \ref{bench}), the evaluation metrics (Section \ref{metrics-sec}), the baseline models (Section \ref{baselines}) and the ablation models (Section \ref{ablations}), each described separately in the following parts.

\subsubsection{The Benchmark for Language-informed Systematic Generalization}
\label{bench}
Here, we explain the benchmark for our experiments. The environment of interest in this work is BabyAI. Since this study focuses on language-informed systematic generalization, we need a language-informed environment in which rich and controllable combinations of subtasks are possible. Compared to other environments described in Section \ref{apprelatedworks}, BabyAI quite satisfies these requirements, and therefore, we choose to \textcolor{black}{evaluate our method} on this environment. BabyAI contains 19 procedurally-generated levels in a grid-world environment. For each level, a set of natural-looking instructions from context-free grammar specify the desired goal. The observations in this environment are mainly partial and symbolic $7\times7\times3$ first-person views. Each entry in a grid cell indicates its entity's type, color, or status, offering a factorized input that makes the learning process much more computationally efficient. This observation space aligns with the \textit{theory of systems 1 \& 2} \citep{booch2021thinking}, separating the entity perception problem from the reasoning required to solve the task. Doing so creates a suitable and logically rich test bed for solely assessing the reasoning ability of the model. 

Given the compositional nature of language, we can define our evaluation protocol, i.e., train/test split of tasks, based on different combinations of possible factors of variation per level, as encouraged by \citep{kirk_survey_2023}. The BabyAI environment does not readily include this separation, and train/test splits are typically created based on random seeds. However, since each seed corresponds to a unique pair of (\textit{initial state}, \textit{instruction}), adding a filter on seeds to store them for specific instructions is a straightforward way to build the systematic split based on the different combination of features, instructions, and entities. The systematic split for each environment is stated in Table \ref{splits-table}. This split is defined based on matching strings inside the instruction; i.e. if the instruction contains any of the specified strings, its seed is going to be reserved for test, otherwise the generated episode is used during training.

The details about the environments are explained below. The environments are chosen to be light-weighted, yet endowed with sufficiently complex logic, regarding the amount of available compute.
We choose fast-converging BabyAI levels, namely ActionObjDoor, GoToSeq, PutNextLocal, PickupLoc, OpenDoorsOrder, and Synth, so that the coverage on different capabilities is considered. 
Each model has been run across two random seeds in the specified environment. The exact name of each level in the BabyAI environment is written inside parentheses.
\begin{itemize}
 \item \textbf{PutNextLocal (\texttt{BabyAI-PutNextLocalS6N4-v0})}: In this level, the agent is instructed to put an object -specified by color and type- next to another object in a single room environment with four objects. Instructions take the form of "\texttt{put the \{color\} \{type\} next to the \{color\} \{type\}}". Color can be "\texttt{red}", "\texttt{blue}", "\texttt{yellow}", "\texttt{green}", "\texttt{grey}", or "\texttt{purple}" and the type can be "\texttt{ball}", "\texttt{key}", or "\texttt{box}". 
 \item \textbf{PickupLoc (\texttt{BabyAI-PickupLoc-v0})}: Instructions in this single-room level take the form of "\texttt{pick up the \{color\} \{type\} \{location\}}" where the color and the type are the same as the previous level, but a location also describes the object of interest -"\texttt{on your left/right}", "\texttt{in front of you}", or "\texttt{behind you}"; for example, "\texttt{Pickup the red box in front of you}".
 \item \textbf{GoToSeq (\texttt{BabyAI-GoToSeqS5R2-v0})}: In this level, the agent is instructed to go to several objects in a specific orders. \textcolor{black}{The instructions\ consists of a variable number of "\texttt{go to a/the \{color\} \{type\}}", "\texttt{and go to a/the \{color\} \{type\}}" and "\texttt{, then go to a/the \{color\} \{type\}}" subtasks}. We use a four-room version of this level where each room's size is $5 \times 5$.
 \item \textbf{ActionObjDoor (\texttt{BabyAI-ActionObjDoor-v0})}: In this single-room level the agent can be instructed to perform multiple verbs such as "\texttt{pick up the \{color\} \{type\}}", "\texttt{go to the \{color\} \{type\}}" or "\texttt{open a \{color\} door}". The colors are the same as the previous environments but the type can also be "\texttt{door}".
 \item \textbf{OpenDoorsOrder (\texttt{BabyAI-OpenDoorsOrderN4-v0})}: This level contains four doors and the agent needs to open some of them in a specific order instructed by a sentence in the this format: "\texttt{open the \{color\} door, the open the \{color\} door}" or "\texttt{open the \{color\} door after you open the \{color\} door}" or "\texttt{open the \{color\} door}".
 \item \textbf{Synth (\texttt{BabyAI-SynthS5R2-v0})}: This level contains "\texttt{pick up a/the \{color\} \{type\}}", "\texttt{go to the \{color\} \{type\}}", "\texttt{open the \{color\} door}", and "\texttt{put the \{color\} \{type\} next to the \{color\} \{type\}}" instructions provided to the agent as a single step task. Similar to GoToSeq, a version with four $5\times 5$ rooms is considered.
\end{itemize}

\begin{table}[!htbp]
 \caption{Evaluation protocol for the selected BabyAI levels based on held-out instructions}
 \label{splits-table}
 \centering
 \begin{tabular}{m{3cm} m{9cm}}
 \toprule
 \textbf{BabyAI Level} & \textbf{Test Split ($G_{test}$)} \\ \midrule
 \texttt{ActionObjDoor} & \multirow{4}{9cm}{Instructions containing these combinations of objects: "\texttt{\textcolor{black}{red box}}", "\texttt{\textcolor{green}{green ball}}", "\texttt{\textcolor{myviolet}{purple key}}", "\texttt{\textcolor{yellow}{yellow box}}", "\texttt{\textcolor{blue}{blue ball}}", "\texttt{\textcolor{gray}{grey key}}"}\\
\texttt{GoToSeq} & \\
\texttt{PutNextLocal} & \\
\texttt{PickupLoc} & \\
\midrule
\texttt{OpenDoorsOrder} & 
Instructions containing these orders of doors:

"\texttt{open the \textcolor{blue}{blue} door, then open the \textcolor{yellow}{yellow} door}",

"\texttt{open the \textcolor{green}{green} door, then open the \textcolor{gray}{grey} door},

"\texttt{open the \textcolor{gray}{grey} door, then open the \textcolor{black}{red} door}",

"\texttt{open the \textcolor{yellow}{yellow} door, then open the \textcolor{myviolet}{purple} door}",

"\texttt{open the \textcolor{black}{red} door, then open the \textcolor{green}{green} door}",

"\texttt{open the \textcolor{myviolet}{purple} door, then open the \textcolor{blue}{blue} door}"\\
\midrule
\texttt{Synth} &
  "\texttt{put the \textcolor{black}{red} ball next to the \textcolor{green}{green} key}",
  
  "\texttt{put the \textcolor{myviolet}{purple} box next to the \textcolor{yellow}{yellow} ball}",
  
  "\texttt{put the \textcolor{blue}{blue} key next to the \textcolor{gray}{grey} box}",

  "\texttt{go to the \textcolor{black}{red} box}",
  
  "\texttt{go to the \textcolor{green}{green} ball}",

  "\texttt{pick up a/the \textcolor{myviolet}{purple} key}",
  
  "\texttt{pick up a/the \textcolor{yellow}{yellow} box}",

  "\texttt{open the \textcolor{blue}{blue} door}",
  
  "\texttt{open the \textcolor{gray}{grey} door}", \\
 \bottomrule
 \end{tabular}
\end{table}

\subsubsection{Evaluation Metrics}
\label{metrics-sec}
In the goal-conditioned settings, it is common to measure the performance of the agent using Success Rate (SR) \citep{liu_goal-conditioned_2022}. Specifically, in the BabyAI tasks, because of the negative effect of the lengthy episodes on magnitude of the final reward, we also use Mean Return (MR) to consider the ability of the agents to avoid unnecessary interactions with the environment. High values of SR and MR on the test split indicate the \textcolor{black}{model's effectiveness} in terms of systematic generalization. Other important measures of out-of-distribution generalization include the Generalization Gap (GG), as proposed in \citep{kirk_survey_2023}, to assess the difference between test time and training time performances, similar to supervised learning. 
\textcolor{black}{We define GG as the amount by which the MR at training time exceeds the MR at test time. Lower values are obviously more desired.}
Another important metric is Sample Efficiency (SE), defined as the minimum number of training frames that the agent needs to see to achieve a certain SR, $\alpha$, and preserve it through the rest of the training process. This metric can be defined based on the training time or test time SRs or even based on the MR. Since this study focuses on the out-of-distribution evaluation setting, we calculate this metric using test time SRs. To assess the SE and MR together, one can use Area Under Curve (AUC) of the MR. We calculate this metrics and report it as AUC-MR as well.

In order to be able to average performance across different environments and report one normalized value for a model, one can reformulate GG and SE so that 1) their value lies in $[0, 1]$ and 2) their higher value is more desirable. We call them normalized GG and SE, denoted by ($\Hat{GG}$) and ($\Tilde{SE}$), respectively, and calculated as below:

\begin{align}
 \hat{GG} &= Average_{e \in E}(\frac{1 - GG_e}{\max_{e' \in E} 1 - GG_{e'}})\\
 \tilde{SE} &= Average_{e \in E}(1 - \frac{SE_e}{F_e})
\end{align}
where $E$ is the set of environments and $F_e$ is the total number of training frames.
Now, we can average these values and report a number between 0 and 1 which is preferred when closer to 1. The other metrics have this property inherently. This behavior alleviates comparison of several metrics across different models. We report these metrics in the radar charts of Fig. \ref{plots-radar}.

\subsubsection{Baselines} 
\label{baselines}
We follow the experimental setup introduced in \citep{chevalier-boisvert_babyai_2018} and train all of the models using PPO \citep{schulman_proximal_2017}. \textcolor{black}{Each model is trained by} Adam optimizer with a learning rate of 1e-4 and $\beta$s equal to 0.9 and 0.999. The gradient is back-propagated through 20 consecutive timesteps generated by the current policy across 16 parallel environment processes. The memory is an LSTM layer with a hidden state size of 1024. We train the models up to 30M frames in PickupLoc, 20M frames in PutNextLocal, 20M frames in GoToSeq, 13M frames in Synth, and 8M frames in ActionObjDoor and OpenDoorsOrder levels\footnote{We used two NVIDIA GeForce GTX 1080 Ti, one TITAN V and two TITAN RTX GPUs over two months for the experiments of this paper.}. At test time, 512 episodes from the test splits, specified in Table \ref{splits-table}, are chosen randomly per level and the average results are reported in Section \ref{results}. This study focuses on \textcolor{black}{ designing} the encoder part of the policy; hence, we choose three models pertaining to different encoder architectures from the literature including CNN-GRU and FiLM-BabyAI \citep{chevalier-boisvert_babyai_2018} along with AttentionFusion \citep{cao2020babyai++} to conduct a fair comparison with different encoding architectures: 
\begin{itemize}
 \item \textbf{CNN-GRU:} This model was proposed along with the BabyAI environment. It processes the observation via a convolutional network and feeds its output to the memory. The memory’s output is concatenated to the representation of the instructions from a GRU network and headed down to the actor-critic networks.

 \item \textbf{FiLM-BabyAI:} In \citep{chevalier-boisvert_babyai_2018}, they also utilize a model with two FiLM controllers [30] to merge the observations encoded via a CNN and the instructions embedded using a GRU. The resulting representation is then fed to the memory and then the actor-critic networks.

 \item \textbf{AttentionFusion:} In this model, cross-attention scores are calculated between the instructions' GRU representations and the observations' CNN representations. Based on these scores, a linear combination of the sentence embeddings is produced and concatenated to the CNN representations, later processed by another convolutional network. The final embedding is headed to the memory and the actor-critic networks afterward. The original paper \citep{chevalier-boisvert_babyai_2018} involves descriptive sentences in the attention process, and the instructive sentences are later incorporated using FiLM layers. As we didn’t have descriptive sentences in this study, we only applied attention to the instruction, eliminating the need for additional FiLM layers.
\end{itemize}

\subsubsection{Ablations}
\label{ablations}
As stated in Section \ref{architecture}, this study augment the NPS with some techniques to enhance its abilities in zero-shot generalization to unseen combinations of task properties. In this section, we describe the variants of our model which incorporate the proposed techniques in different parts of the network. Note that in all variants, we have used the parallel version of the NPS in which we select one rule per slot simultaneously. This choice is done since every slot in the observation changes at each timestep due to the partial observability of the environment. Moreover, we found that using only one primary slot per time step for the whole observation drastically reduces the performance and for all slots we select and apply rules on them. The proposed techniques fall into the following categories.

\begin{itemize}
 \item \textit{\textcolor{black}{Language Entrance}:} We try the following variants to explore the places where the instruction can enter the model (as mentioned in Section \ref{architecture})\textcolor{black}{, such as late concatenation to middle representations of the observation, and early fusion with observation prior to applying the NPS}.
 \begin{itemize}
 \item \textbf{IC-AC (Instruction-conditioned Actor-Critic):} In this version, the observations are processed using an NPS. The GRU representations of the instructions are concatenated to the representations of the observations as the input of the actor-critic networks.
 \item \textbf{IC-M (Instruction-conditioned Memory):} This variant is similar to the previous one, IC-AC, but the instruction representations are concatenated to the input of the memory instead of actor-critic. 
 \item \textbf{IC-Input (Instruction-conditioned Input):} Using a FiLM controller, we first perform an early fusion of the observation and the instruction at each time step, and the resulting representation passes through the NPS.
 \end{itemize}
 \item \textit{Memory Feedback:} To incorporate the hidden state of the memory in the attention queries, a linear network converts the hidden state of the LSTM to a representation of the query’s size. Then, this representation is concatenated to the query during the rule selection (\textbf{FR}) or contextual slot selection (\textbf{FC}).
\end{itemize}
We also try a \textbf{Raw} model in which the observation is passed to the memory -with a consistent hidden state size- without any layers in between. \textcolor{black}{The instruction representation is concatenated to the actor-critic's input.} This baseline examines the necessity of an observation processing network. 
We discuss these results more in Sections \ref{results} and \ref{sec:disc}.

\subsection{Results}
\label{results}
\begin{figure}[!htbp]

 \begin{subfigure}{\textwidth}
 \centering
 \includegraphics[width=0.6\linewidth]{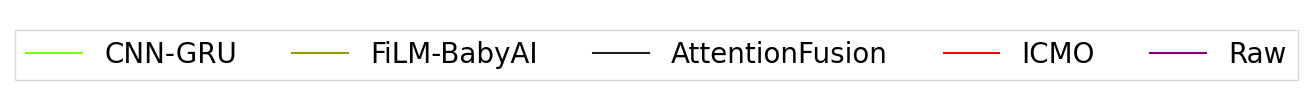}
 \end{subfigure}
 \vspace{0.15em}

 \begin{subfigure}{0.32\textwidth}
 \includegraphics[width=\linewidth]{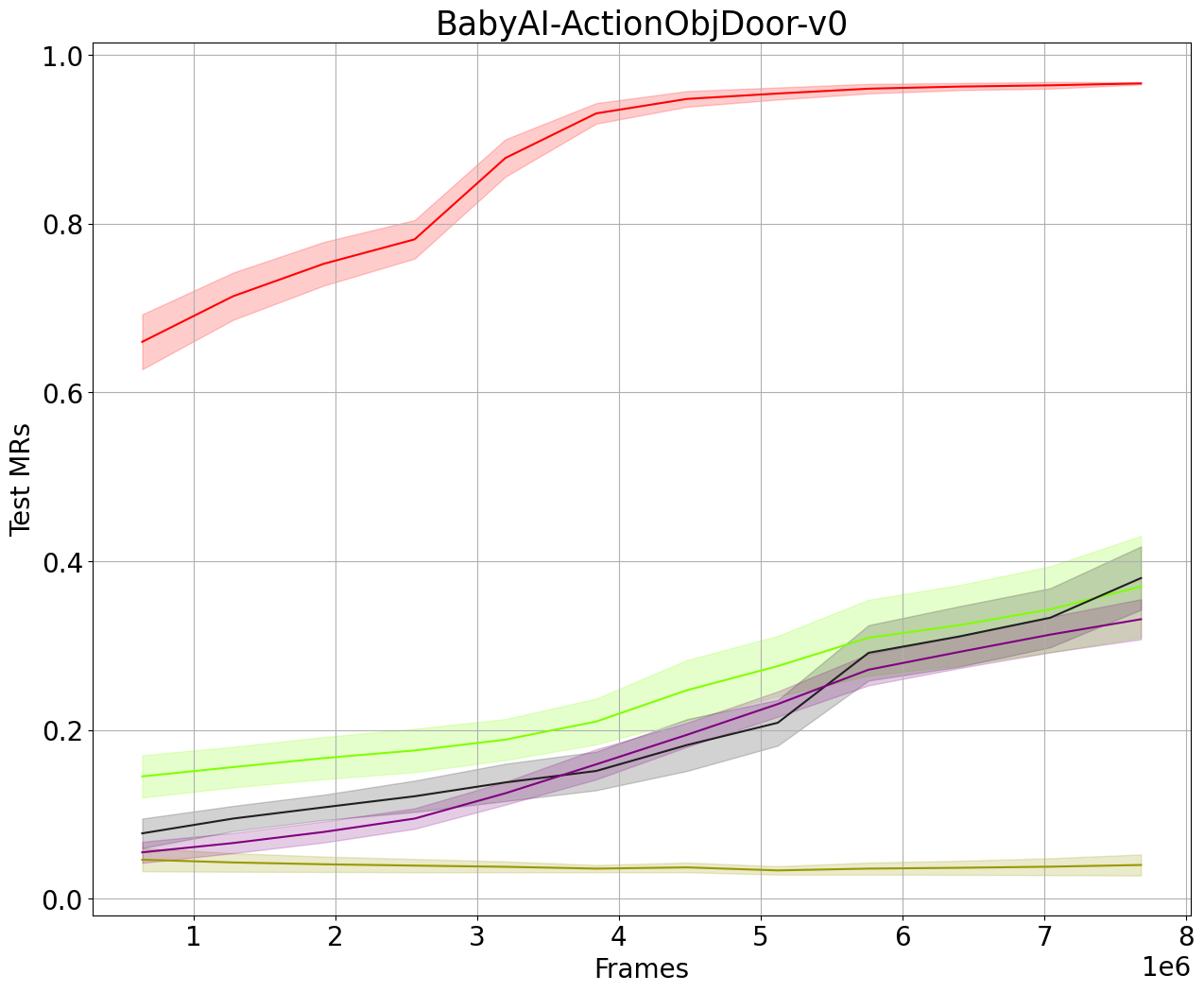}
 \caption{}
 \label{plots-a}
 \end{subfigure}%
 \hfill
 \begin{subfigure}{0.32\textwidth}
 \includegraphics[width=\linewidth]{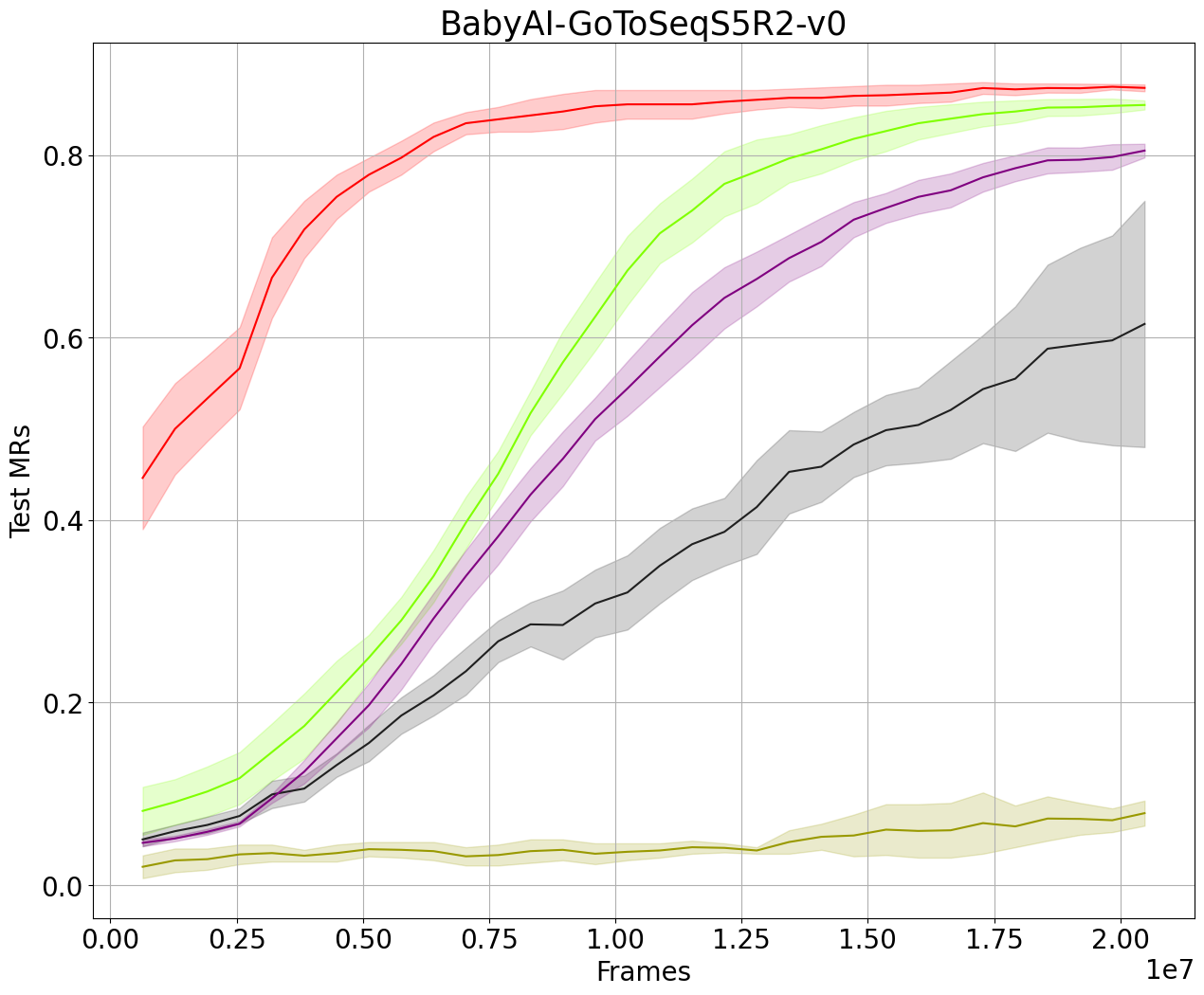}
 \caption{}
 \label{plots-b}
 \end{subfigure}%
 \hfill
 \begin{subfigure}{0.32\textwidth}
 \includegraphics[width=\linewidth]{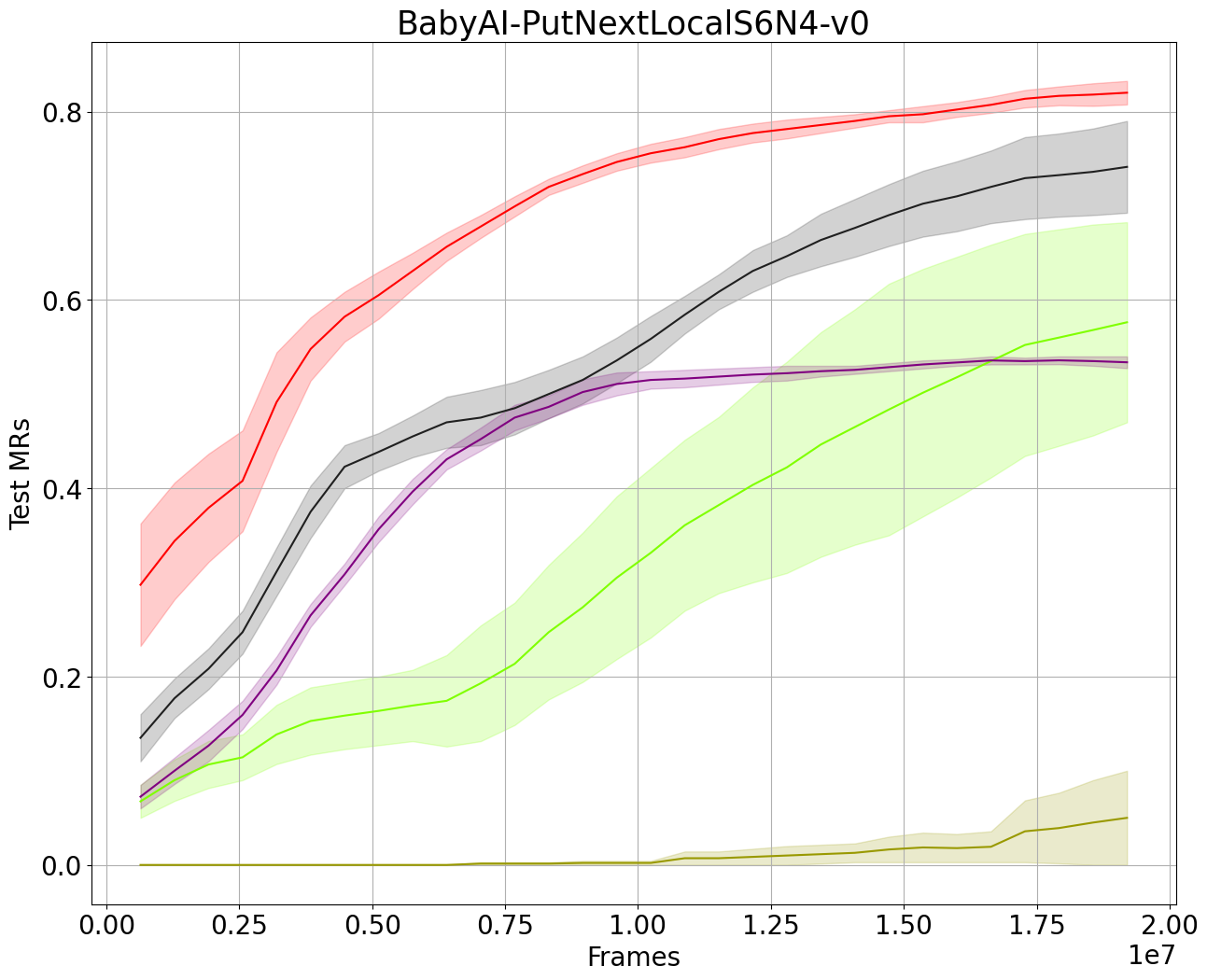}
 \caption{}
 \label{plota-c}
 \end{subfigure}
 
 \begin{subfigure}{0.32\textwidth}
 \includegraphics[width=\linewidth]{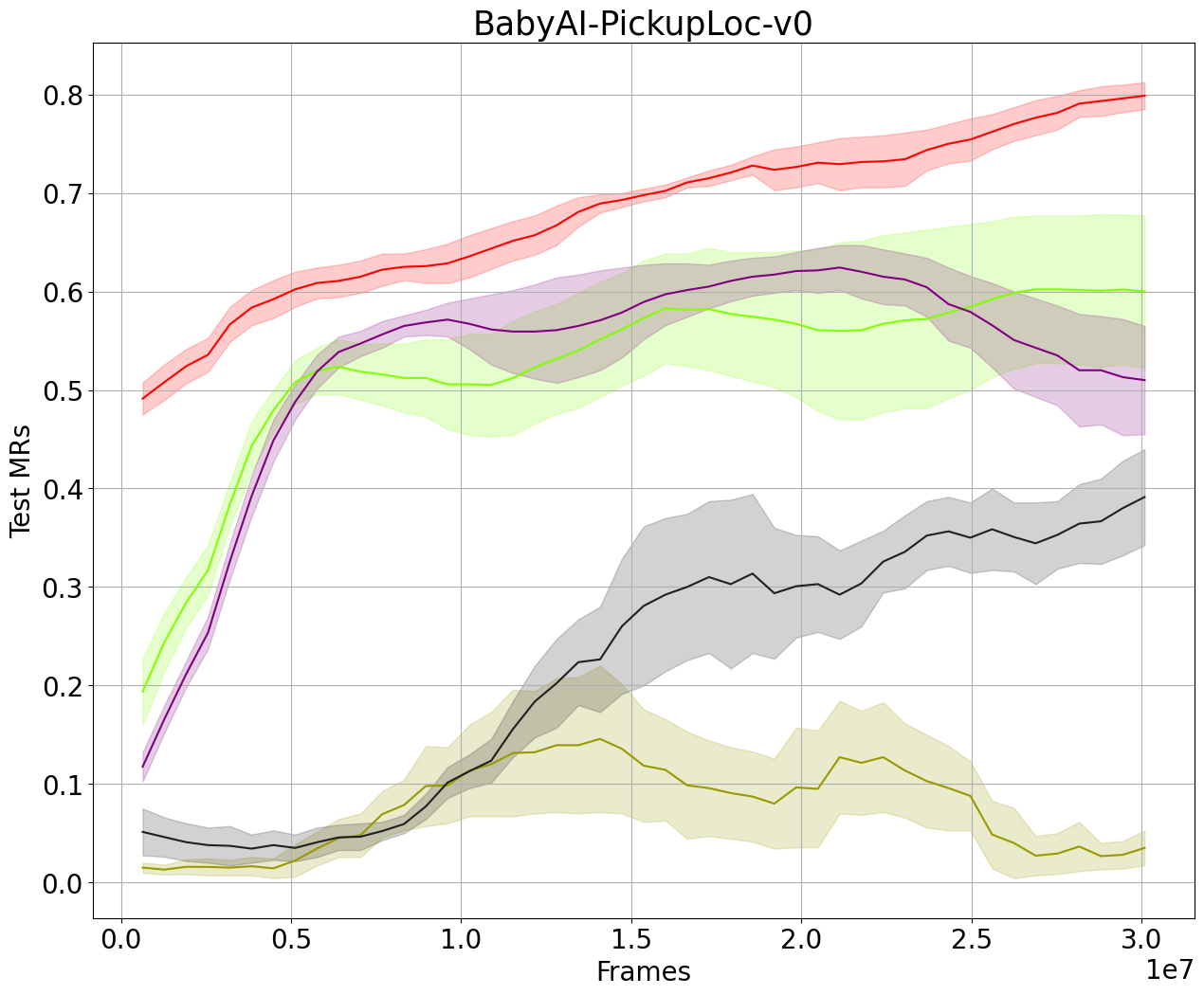}
 \caption{}
 \label{plots-d}
 \end{subfigure}
 \hfill
 \begin{subfigure}{0.32\textwidth}
 \includegraphics[width=\linewidth]{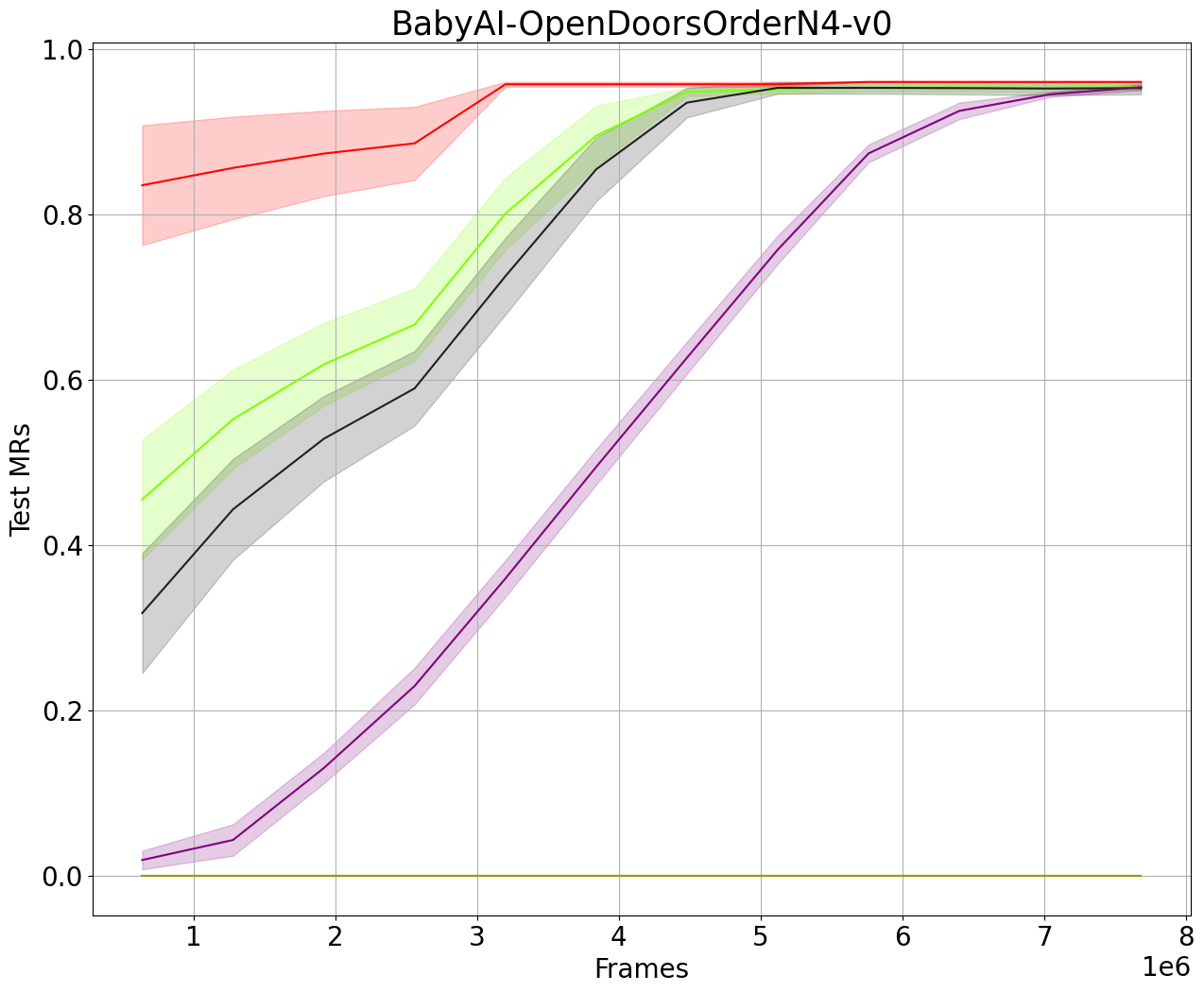}
 \caption{}
 \label{plots-e}
 \end{subfigure}%
 \hfill
 \begin{subfigure}{0.32\textwidth}
 \includegraphics[width=\linewidth]{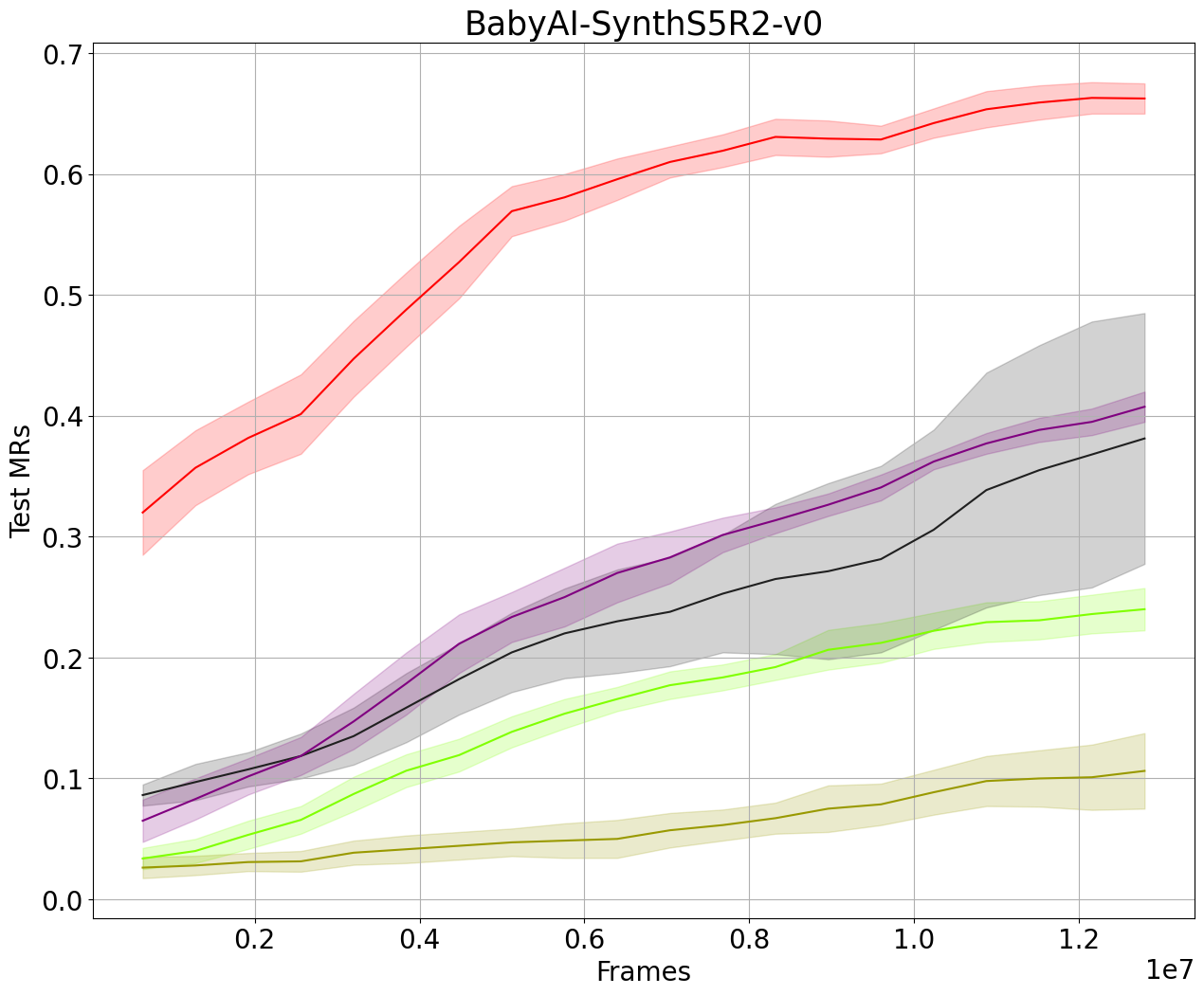}
 \caption{}
 \label{plots-f}
 \end{subfigure}
 
 \begin{subfigure}{0.32\textwidth}
 \includegraphics[width=\linewidth]{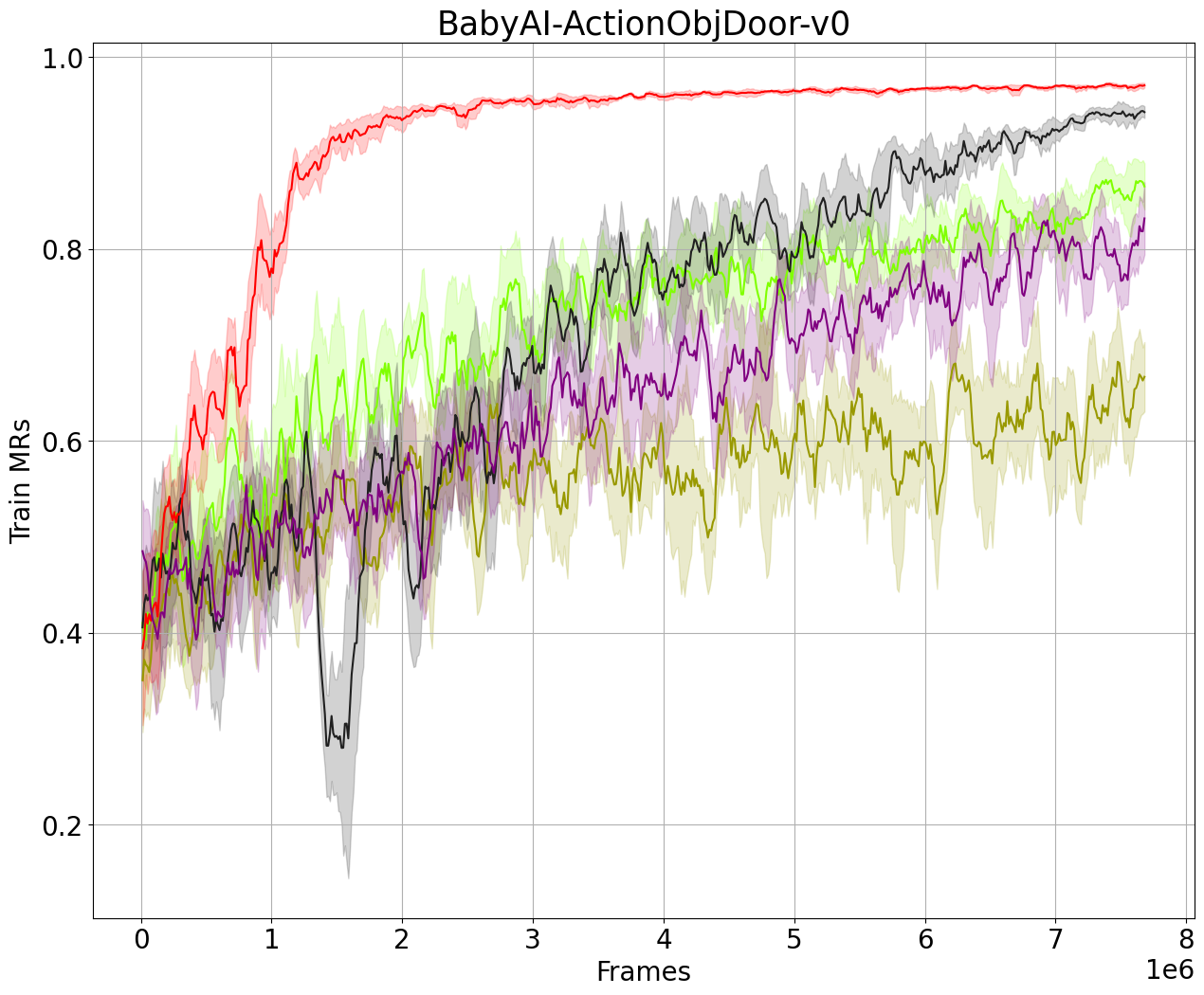}
 \caption{}
 \label{plots-g}
 \end{subfigure}
 \hfill
 \begin{subfigure}{0.32\textwidth}
 \includegraphics[width=\linewidth]{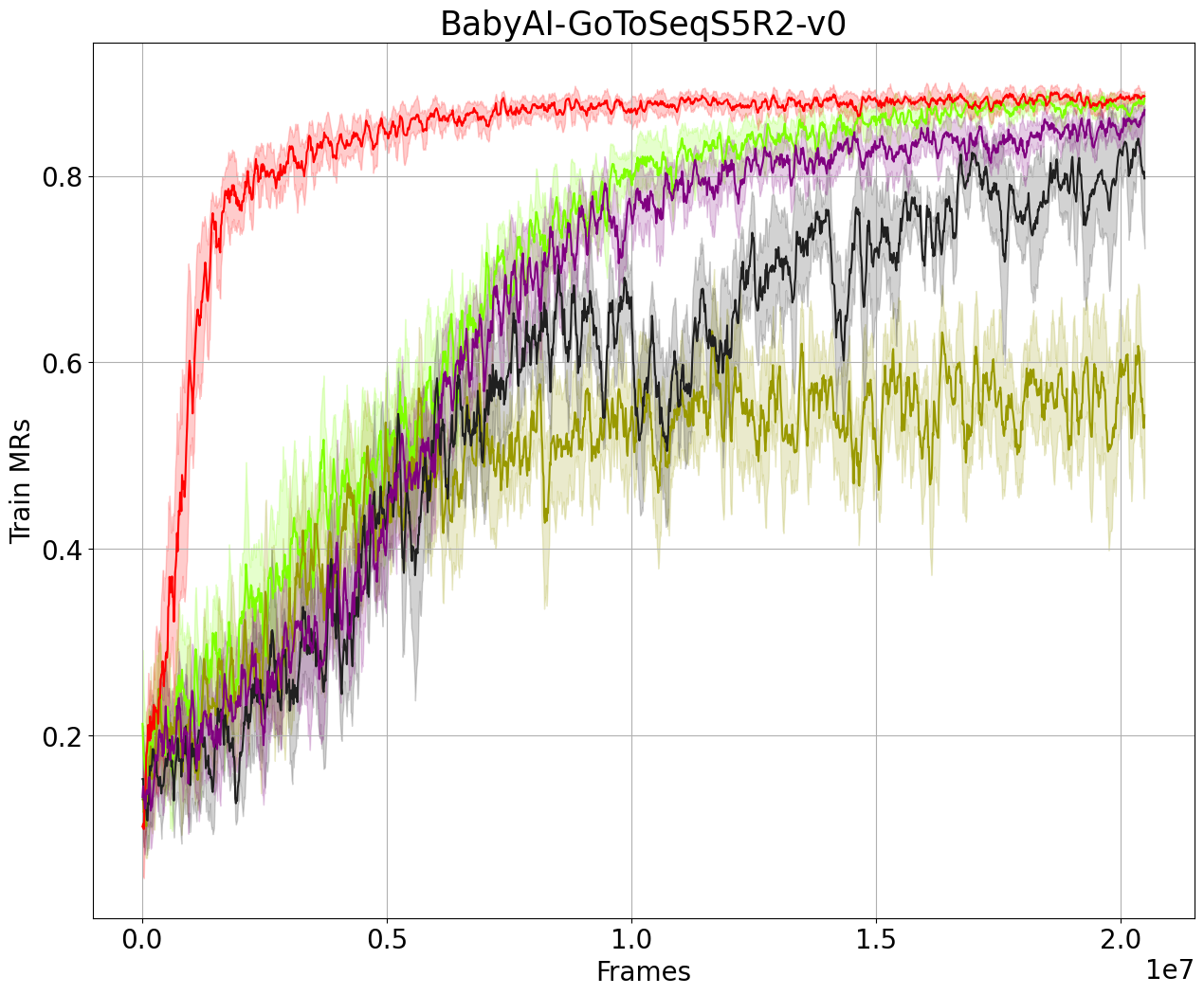}
 \caption{}
 \label{plots-h}
 \end{subfigure}%
 \hfill
 \begin{subfigure}{0.32\textwidth}
 \includegraphics[width=\linewidth]{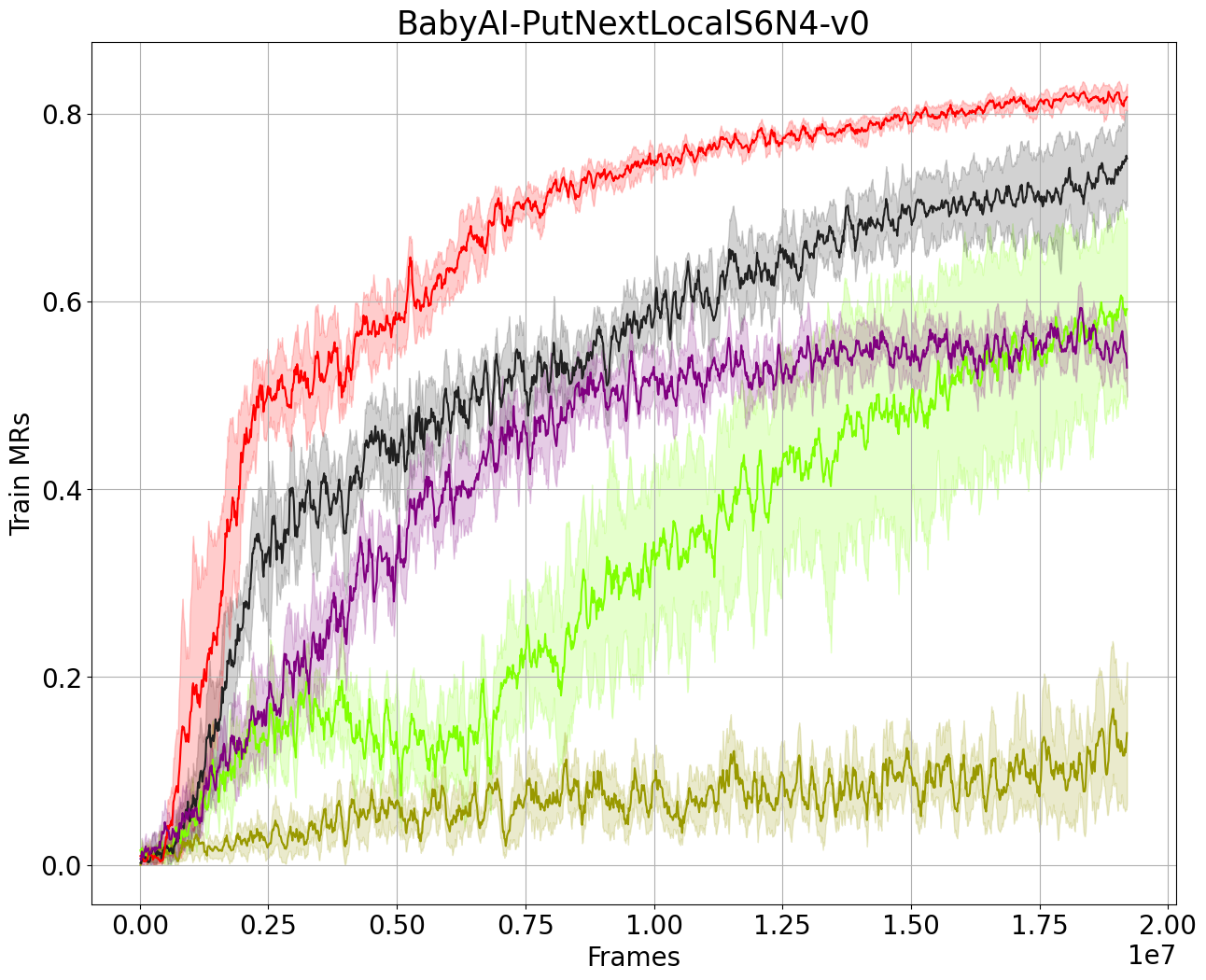}
 \caption{}
 \label{plots-i}
 \end{subfigure}
 
 \begin{subfigure}{0.32\textwidth}
 \includegraphics[width=\linewidth]{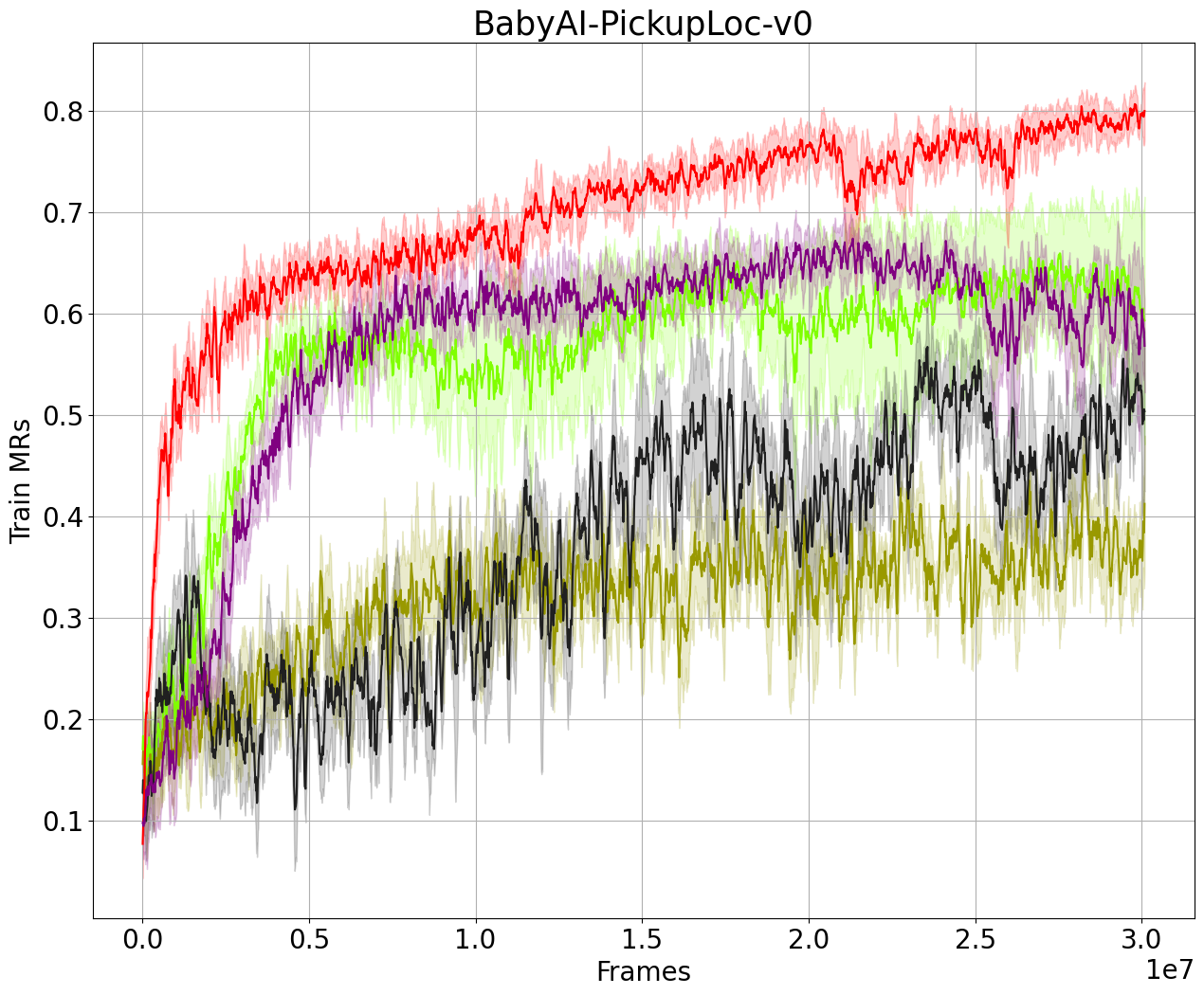}
 \caption{}
 \label{plots-k}
 \end{subfigure}
 \hfill
 \begin{subfigure}{0.32\textwidth}
 \includegraphics[width=\linewidth]{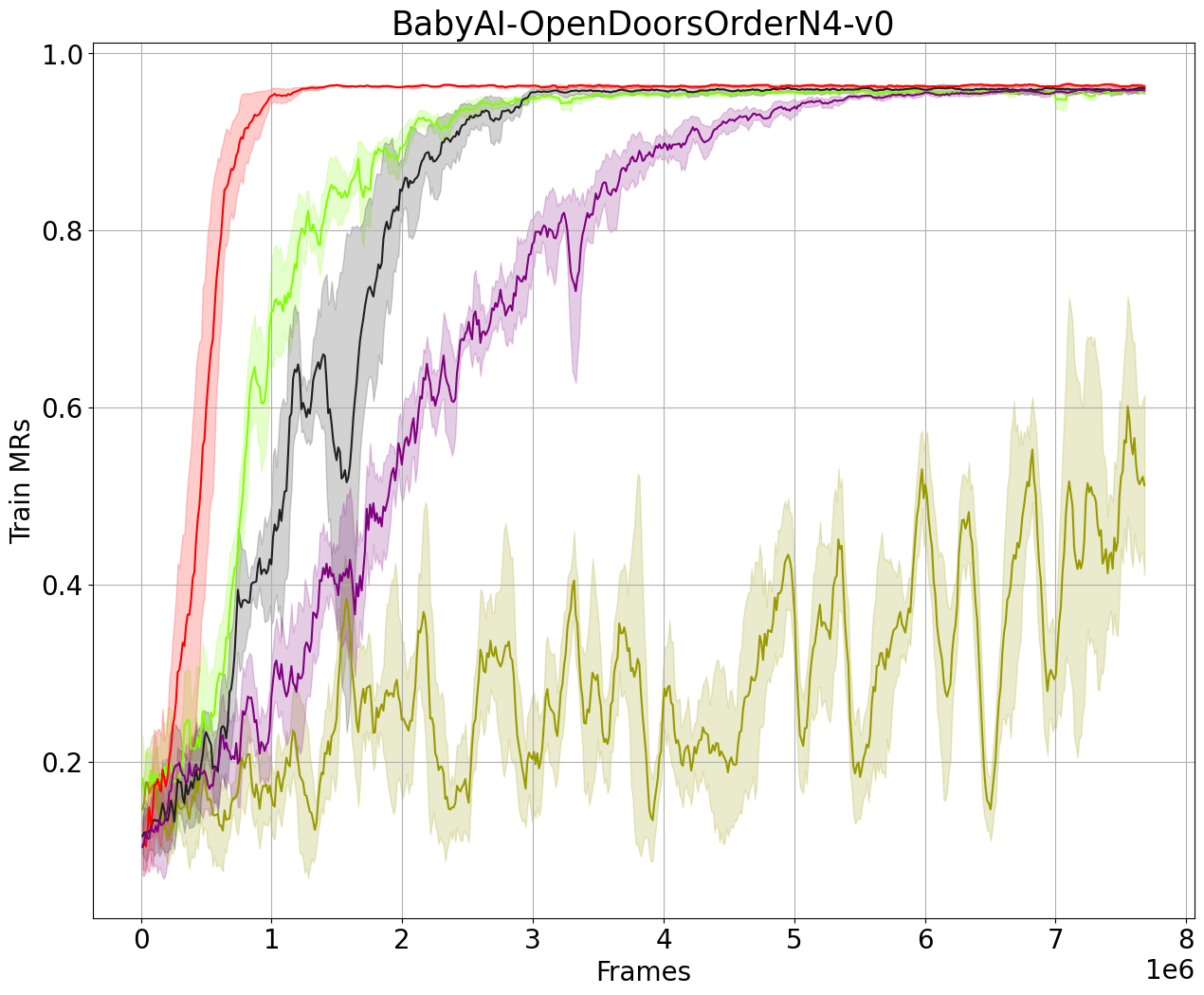}
 \caption{}
 \label{plots-k}
 \end{subfigure}%
 \hfill
 \begin{subfigure}{0.32\textwidth}
 \includegraphics[width=\linewidth]{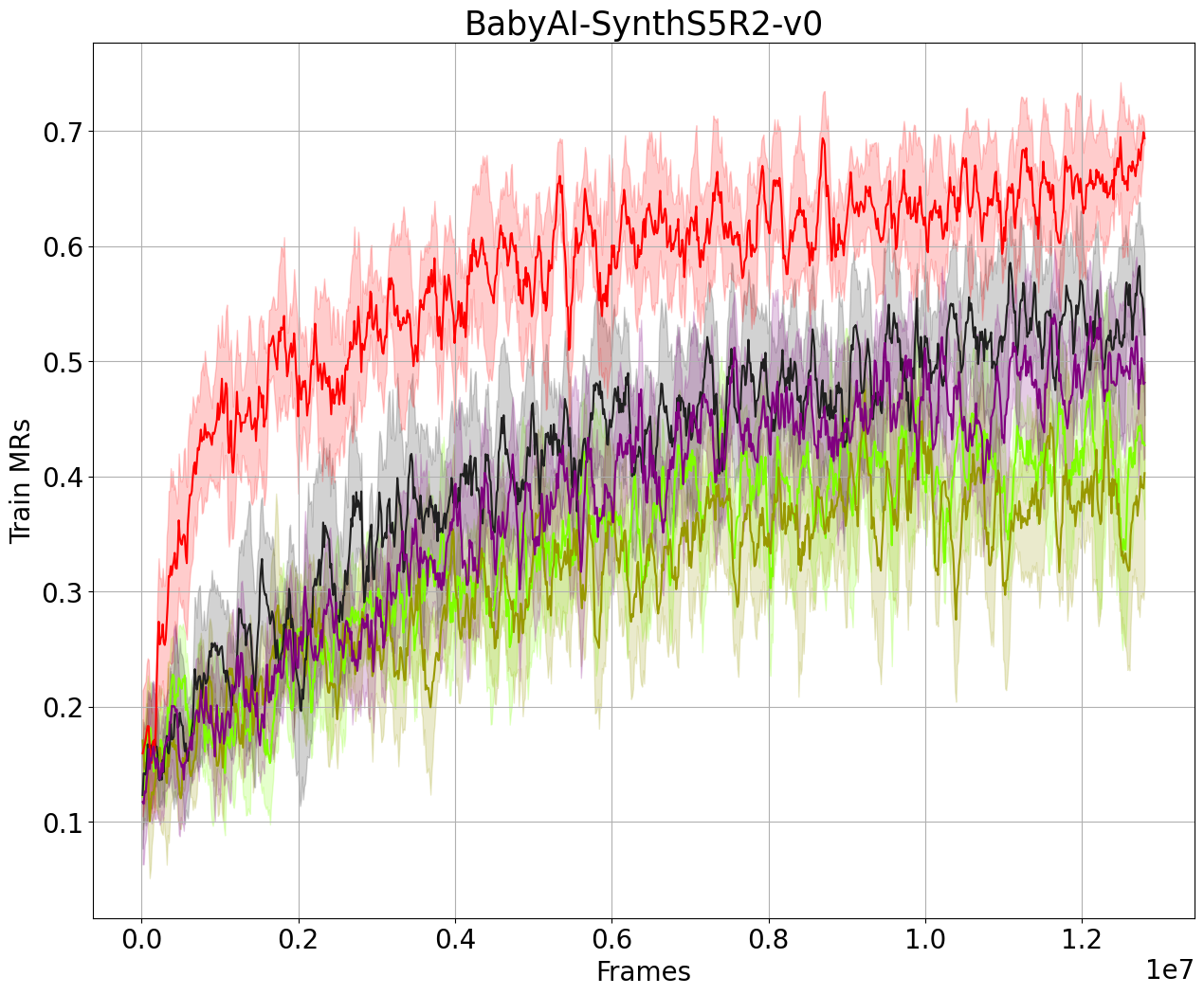}
 \caption{}
 \label{plots-m}
 \end{subfigure}
 
 \caption{Test and train MR trends comparing ICMO to baselines in terms of test MR (a-f) and train MR (g-l)}
 \label{plots}
\end{figure}

In this section, we report the results of the baseline models described in the above subsection. Table \ref{table-baselines} compares the proposed model, ICMO, with the previous models in test SR, test MR, GG, SE($\alpha=0.9$), and AUC-MR. 
The learning curves for train and test MRs are also reported in Fig. \ref{plots}. \textcolor{black}{The training curves indicate performance over $G_{train}$ and test curves are obtained on $G_{test}$ stated for each level in Table \ref{splits-table}. Although this paper focuses on systematic generalization performance and Tables \ref{table-baselines} to \ref{table-memory} report metrics over $G_{test}$, we plot training curves to compare in-distribution performances and showcase the performance gap of models between train/test splits.}
These results indicate that our model outperforms the baselines with a significant margin.

\begin{figure}[!htbp]

 \begin{subfigure}{\textwidth}
 \centering
 \includegraphics[width=0.3\linewidth]{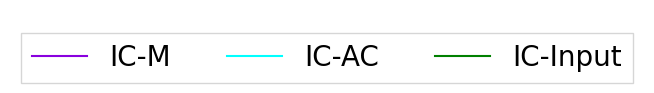}
 \end{subfigure}
 \vspace{0.1em}

 \begin{subfigure}{0.32\textwidth}
 \centering
 \includegraphics[width=\linewidth]{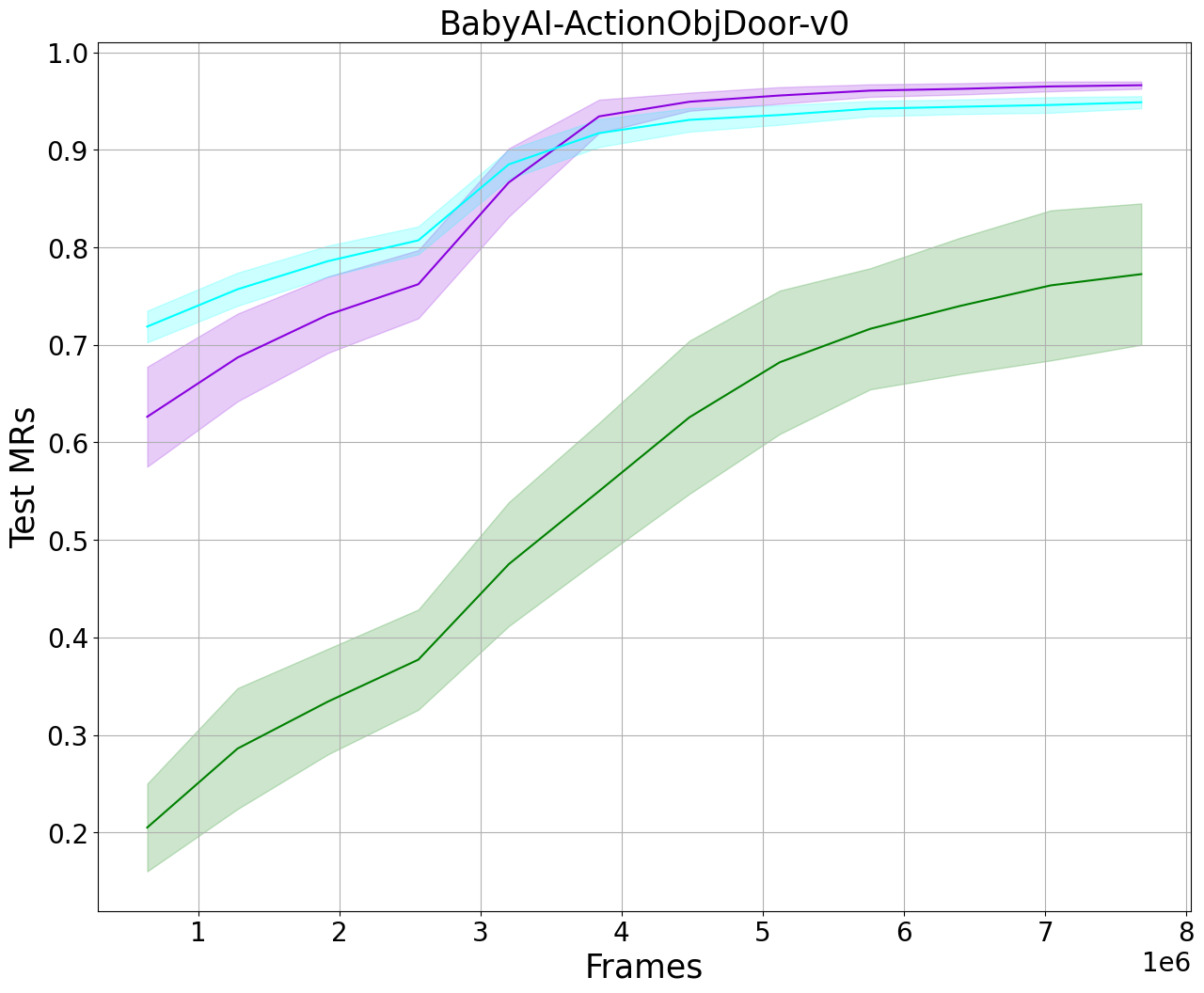}
 \caption{}
 \label{plots-instr-a}
 \end{subfigure}%
 \hfill
 \begin{subfigure}{0.32\textwidth}
 \centering
 \includegraphics[width=\linewidth]{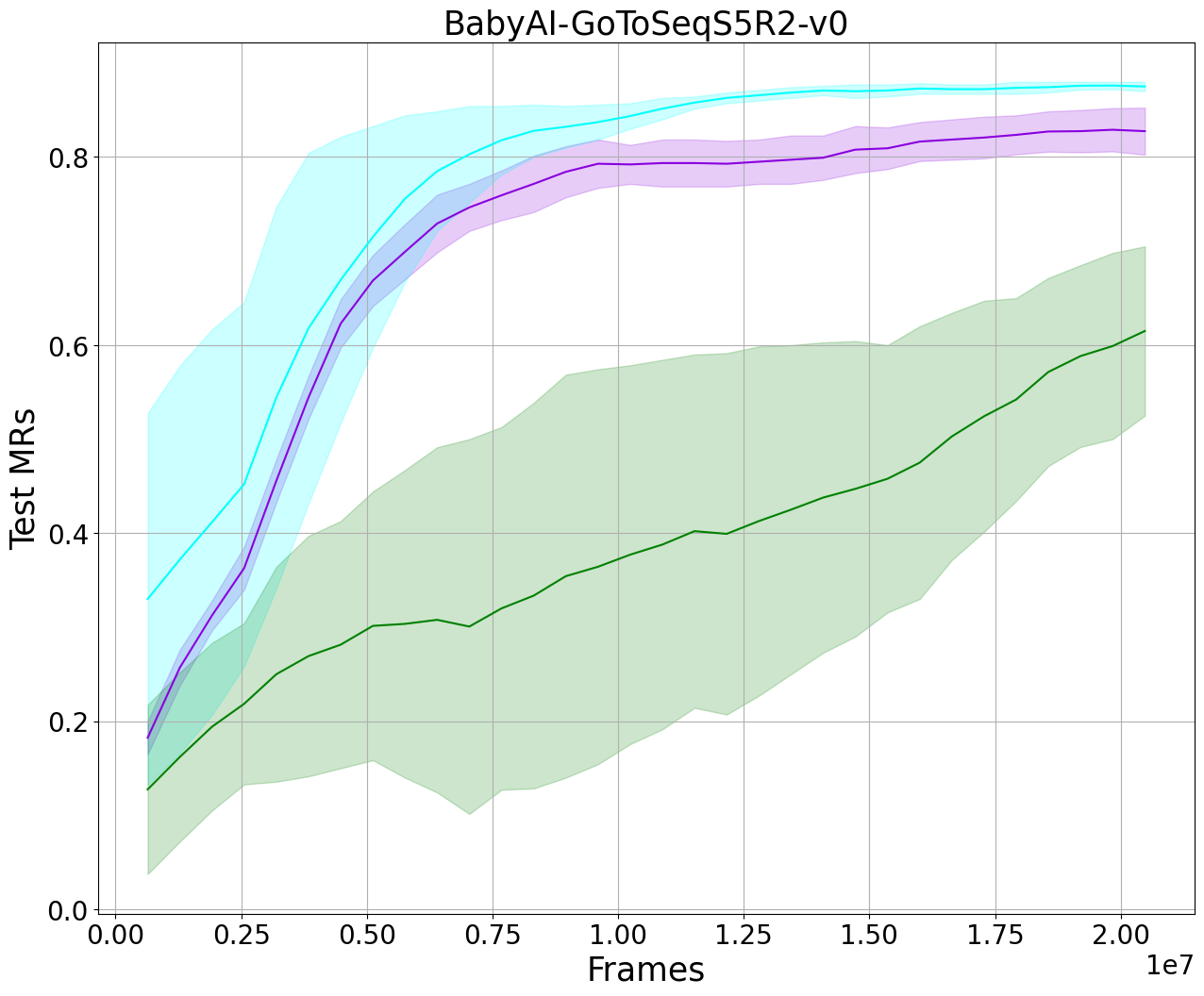}
 \caption{}
 \label{plots-instr-b}
 \end{subfigure}%
 \hfill
 \begin{subfigure}{0.32\textwidth}
 \centering
 \includegraphics[width=\linewidth]{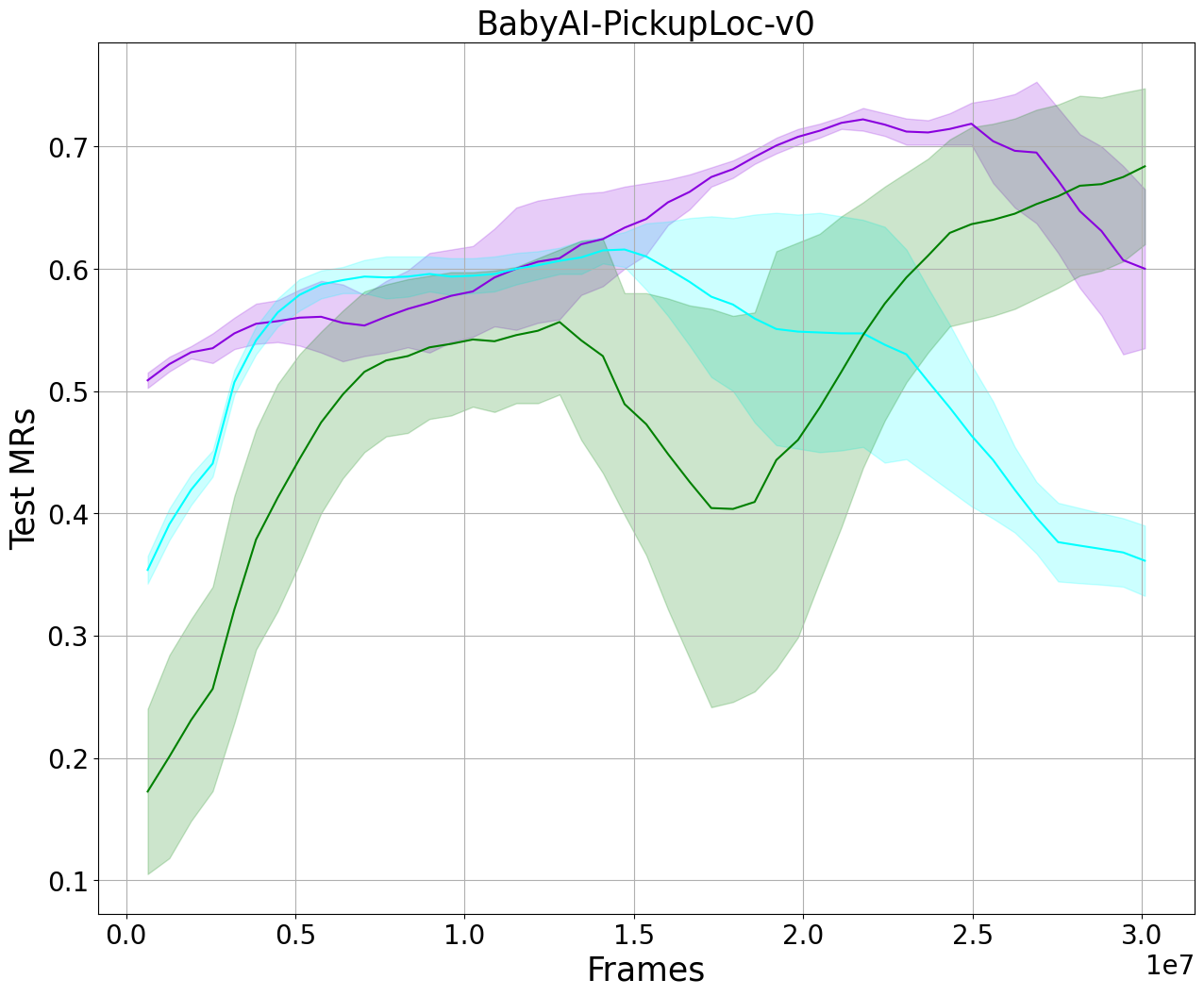}
 \caption{}
 \label{plots-instr-c}
 \end{subfigure}%
 
 \begin{subfigure}{0.32\textwidth}
 \centering
 \includegraphics[width=\linewidth]{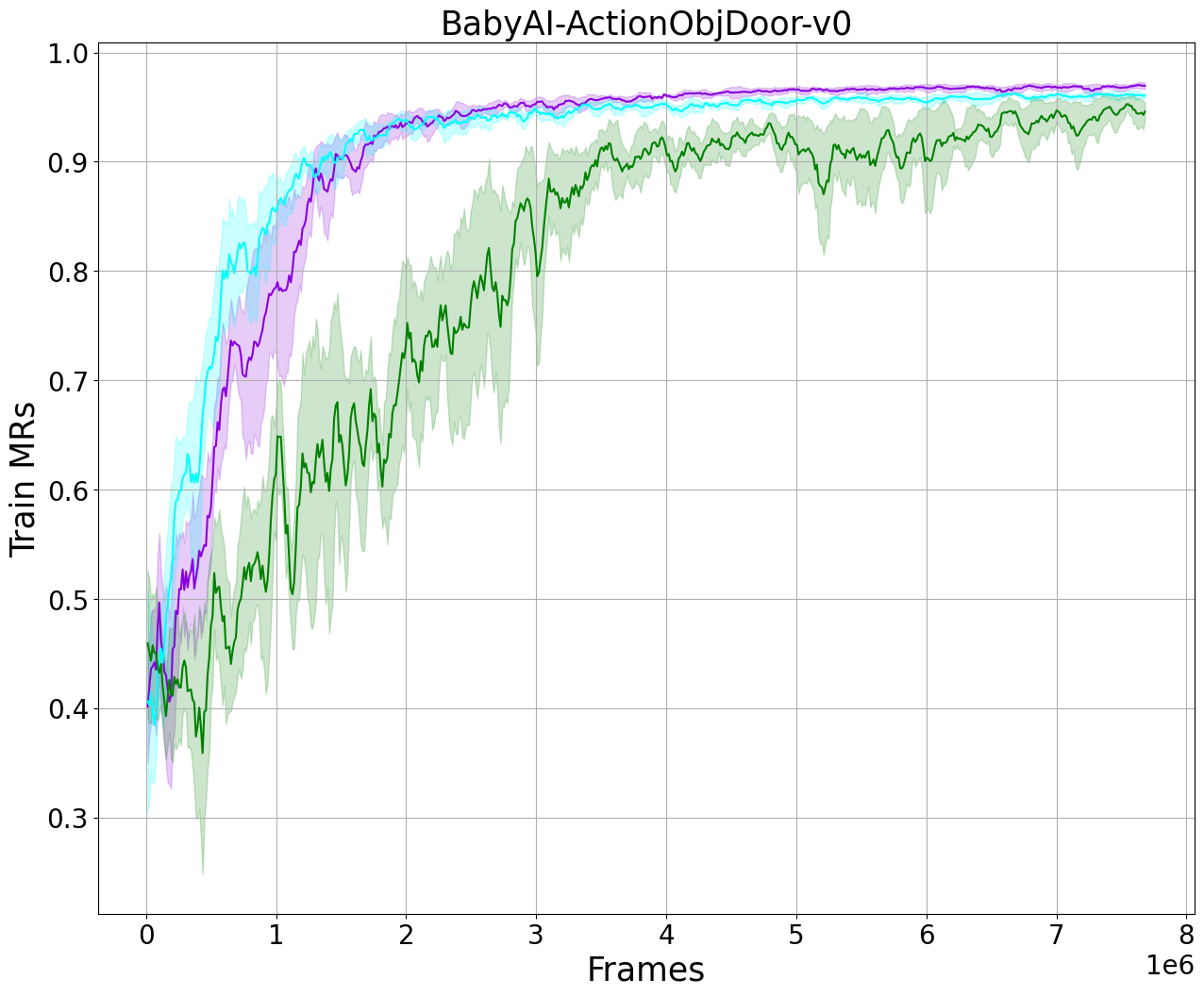}
 \caption{}
 \label{plots-instr-d}
 \end{subfigure}%
 \hfill
 \begin{subfigure}{0.32\textwidth}
 \centering
 \includegraphics[width=\linewidth]{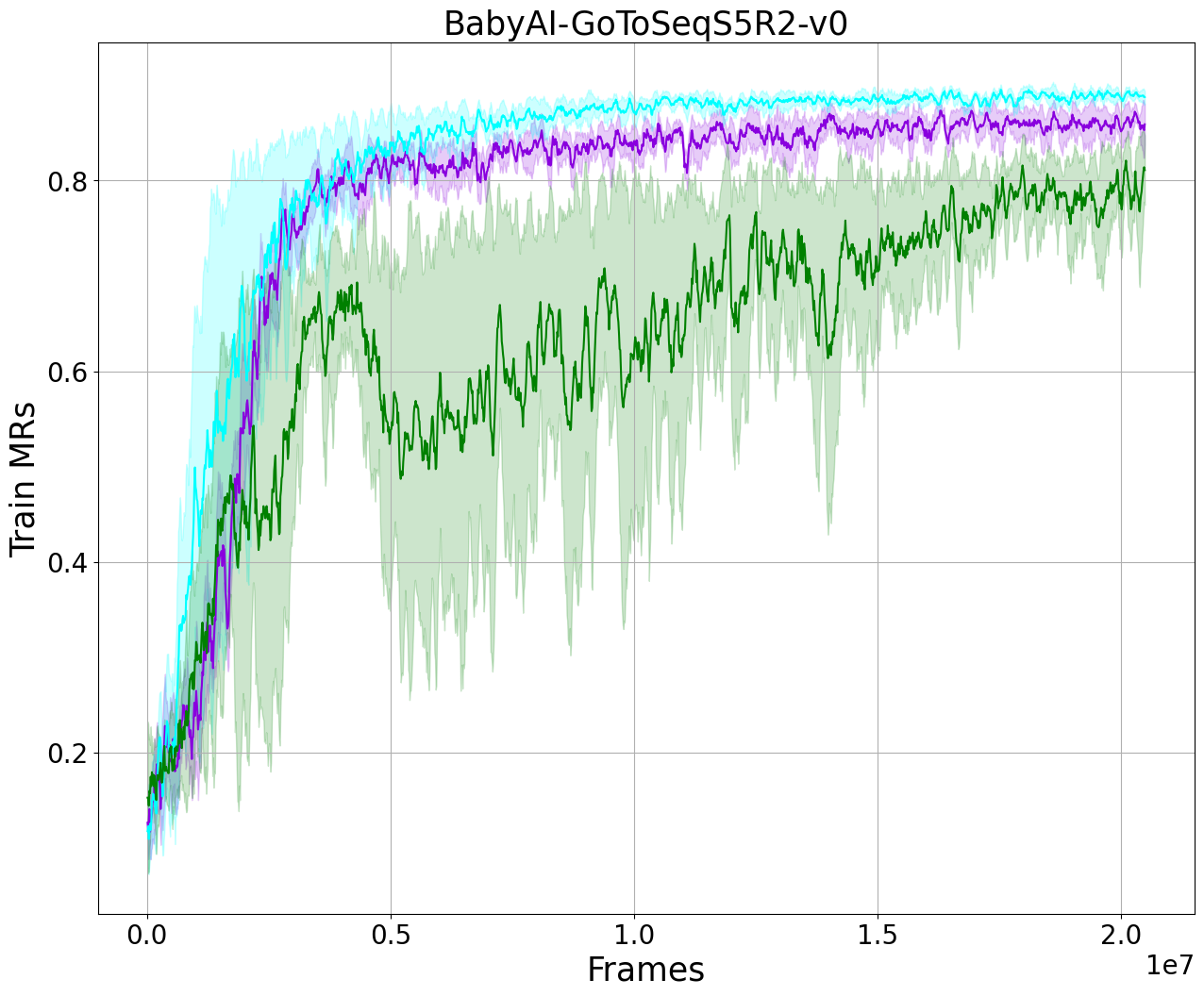}
 \caption{}
 \label{plots-instr-e}
 \end{subfigure}%
 \hfill
 \begin{subfigure}{0.32\textwidth}
 \centering
 \includegraphics[width=\linewidth]{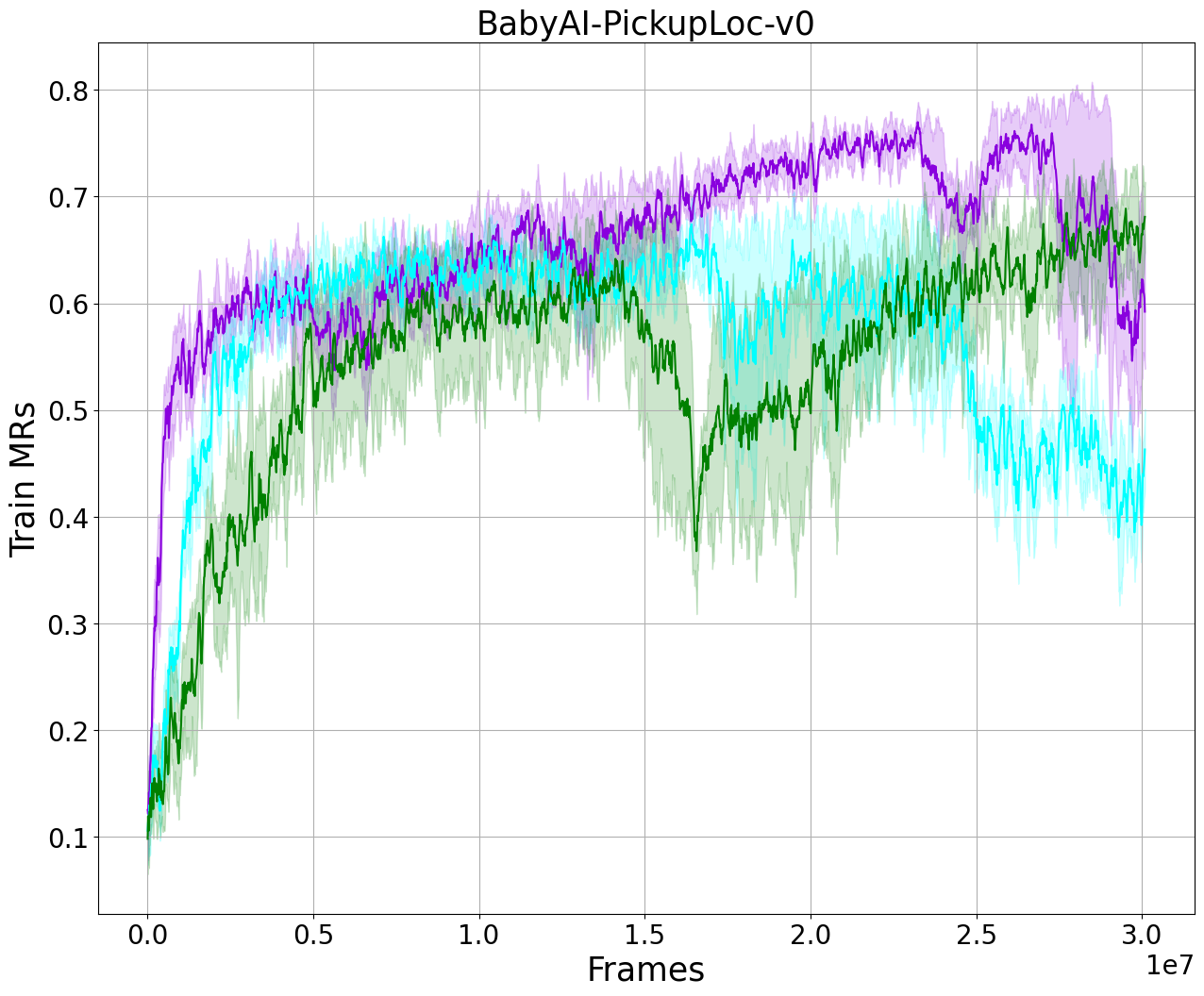}
 \caption{}
 \label{plots-instr-f}
 \end{subfigure}%
 
 \caption{Test and train MR trends comparing instruction ablations}
 \label{plots-instr}
\end{figure}

\begin{figure}[!htbp]

 \begin{subfigure}{\textwidth}
 \centering
 \includegraphics[width=0.6\linewidth]{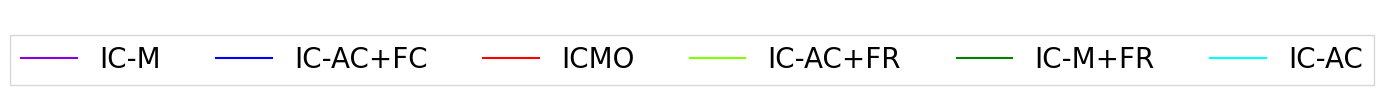}
 \end{subfigure}
 \vspace{0.1em}

 \begin{subfigure}{0.32\textwidth}
 \centering
 \includegraphics[width=\linewidth]{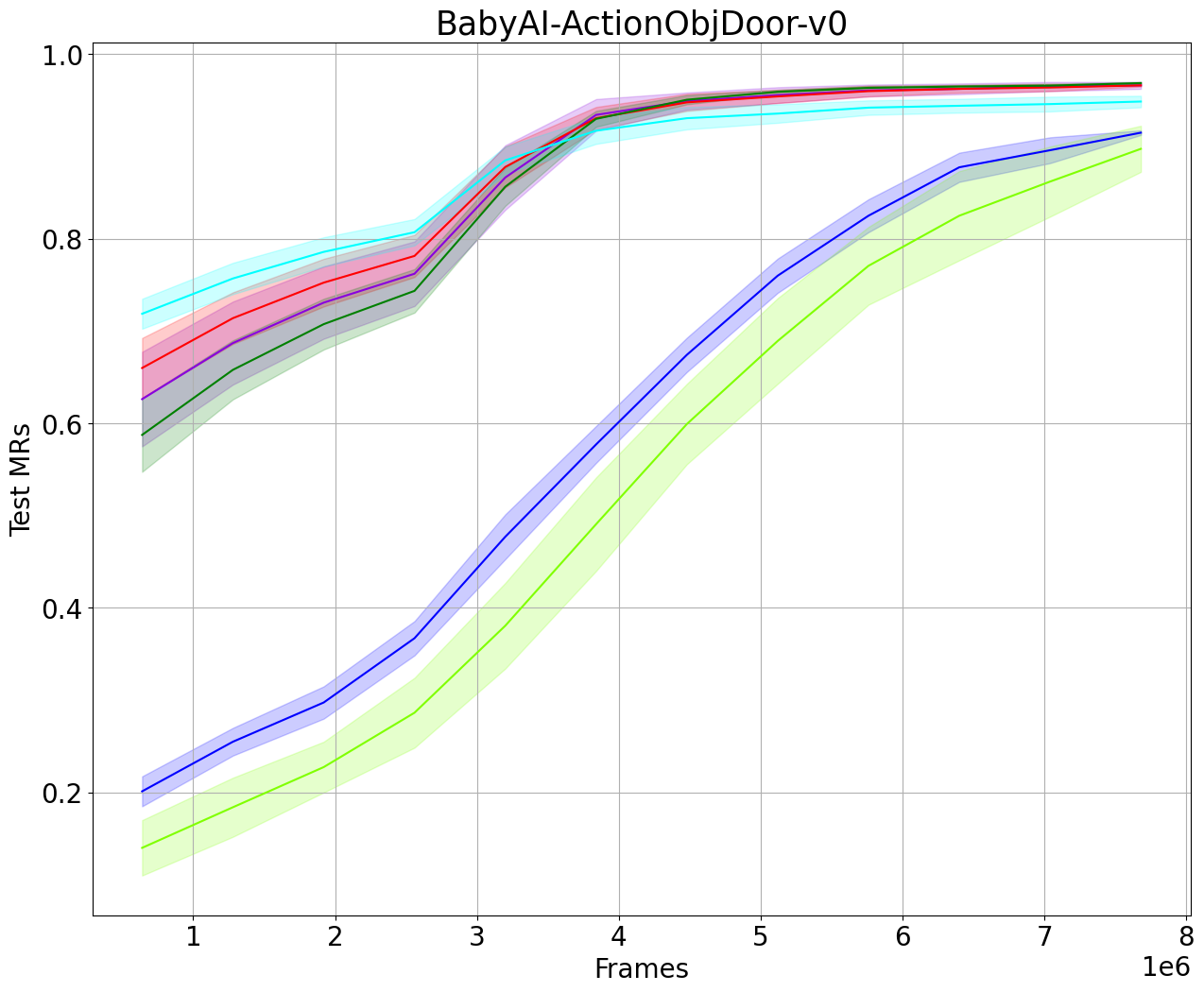}
 \caption{}
 \label{plots-memo-a}
 \end{subfigure}%
 \hfill
 \begin{subfigure}{0.32\textwidth}
 \centering
 \includegraphics[width=\linewidth]{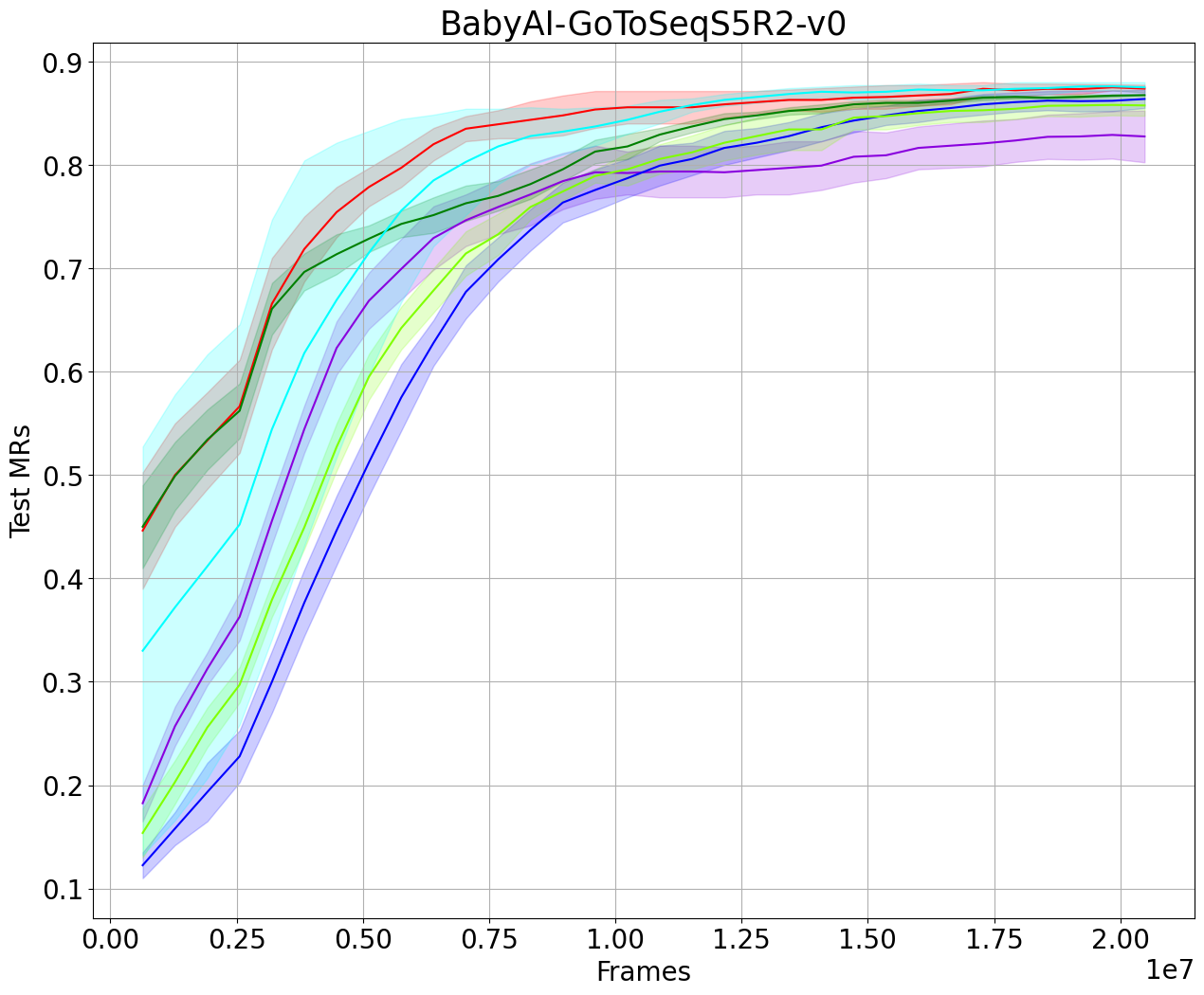}
 \caption{}
 \label{plots-memo-b}
 \end{subfigure}%
 \hfill
 \begin{subfigure}{0.32\textwidth}
 \centering
 \includegraphics[width=\linewidth]{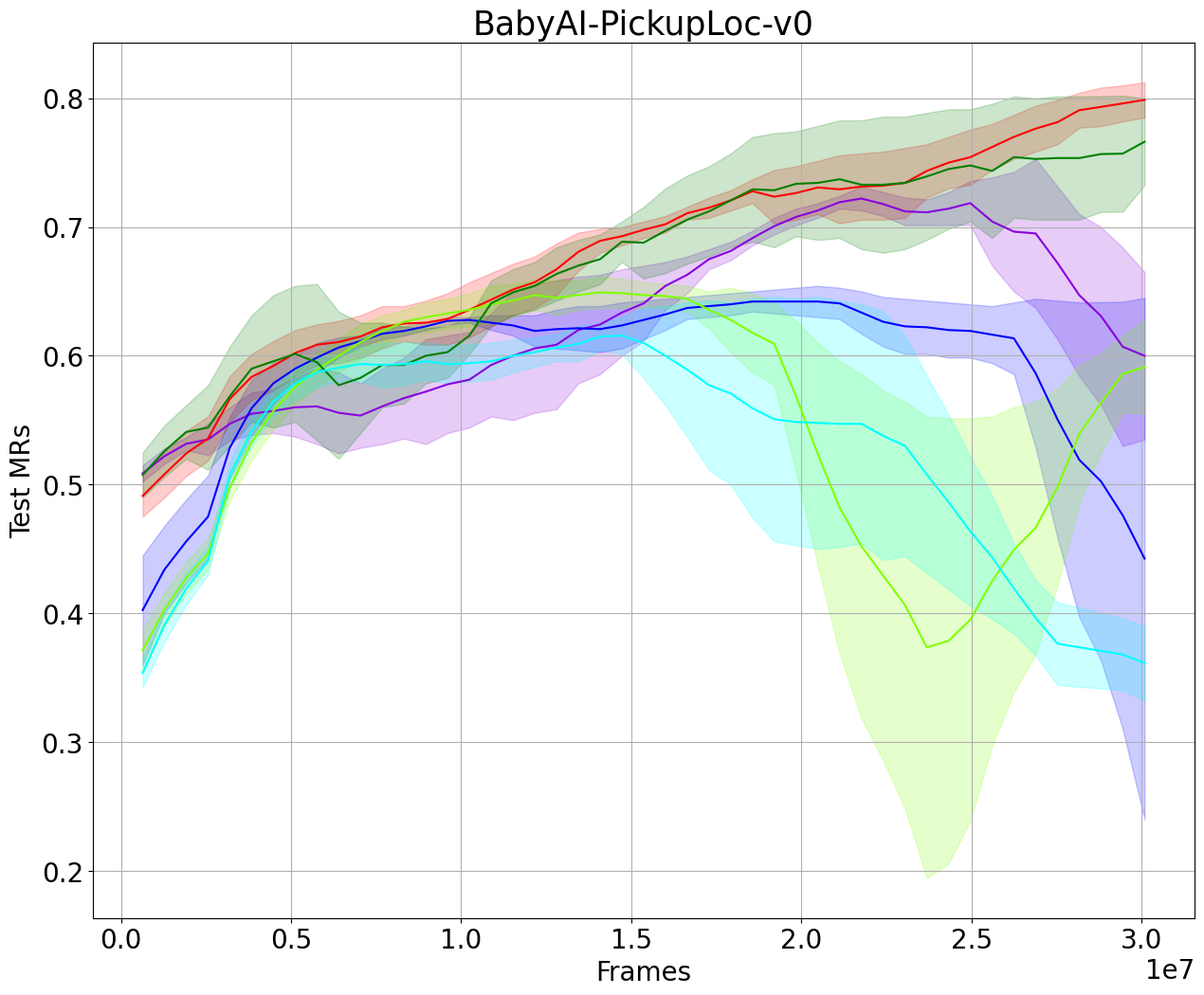}
 \caption{}
 \label{plots-memo-c}
 \end{subfigure}%
 
 \begin{subfigure}{0.32\textwidth}
 \centering
 \includegraphics[width=\linewidth]{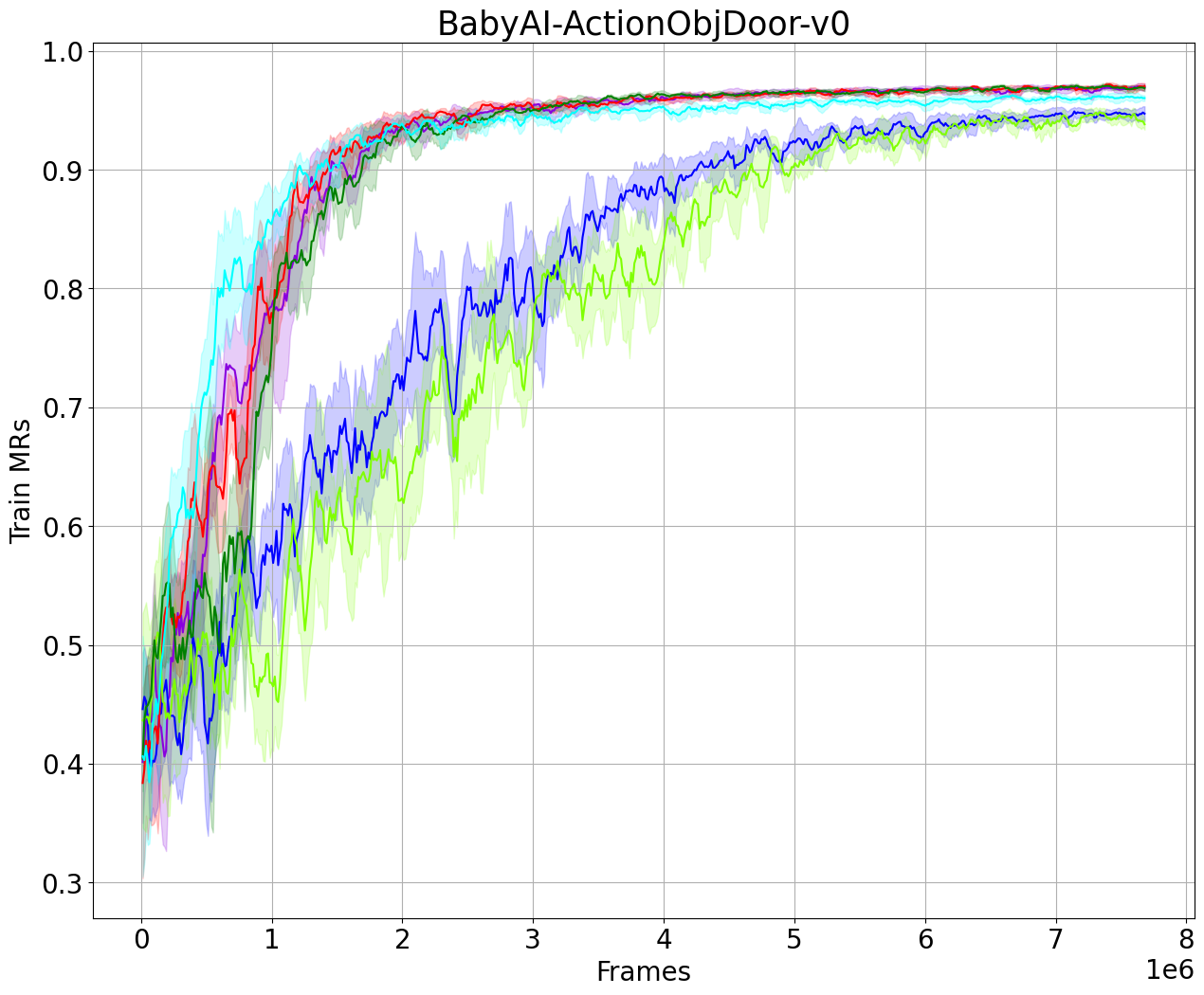}
 \caption{}
 \label{plots-memo-d}
 \end{subfigure}%
 \hfill
 \begin{subfigure}{0.32\textwidth}
 \centering
 \includegraphics[width=\linewidth]{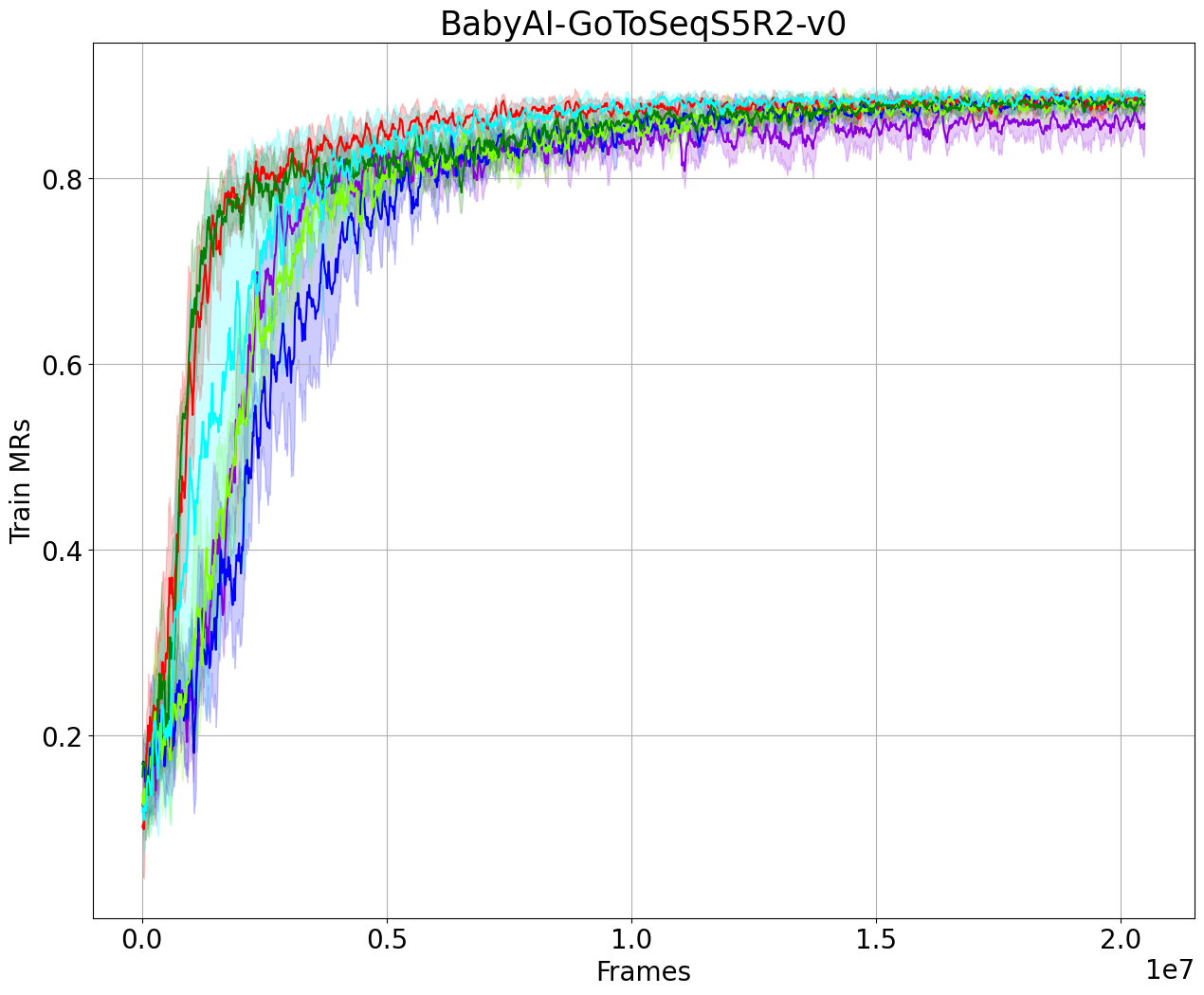}
 \caption{}
 \label{plots-memo-e}
 \end{subfigure}%
 \hfill
 \begin{subfigure}{0.32\textwidth}
 \centering
 \includegraphics[width=\linewidth]{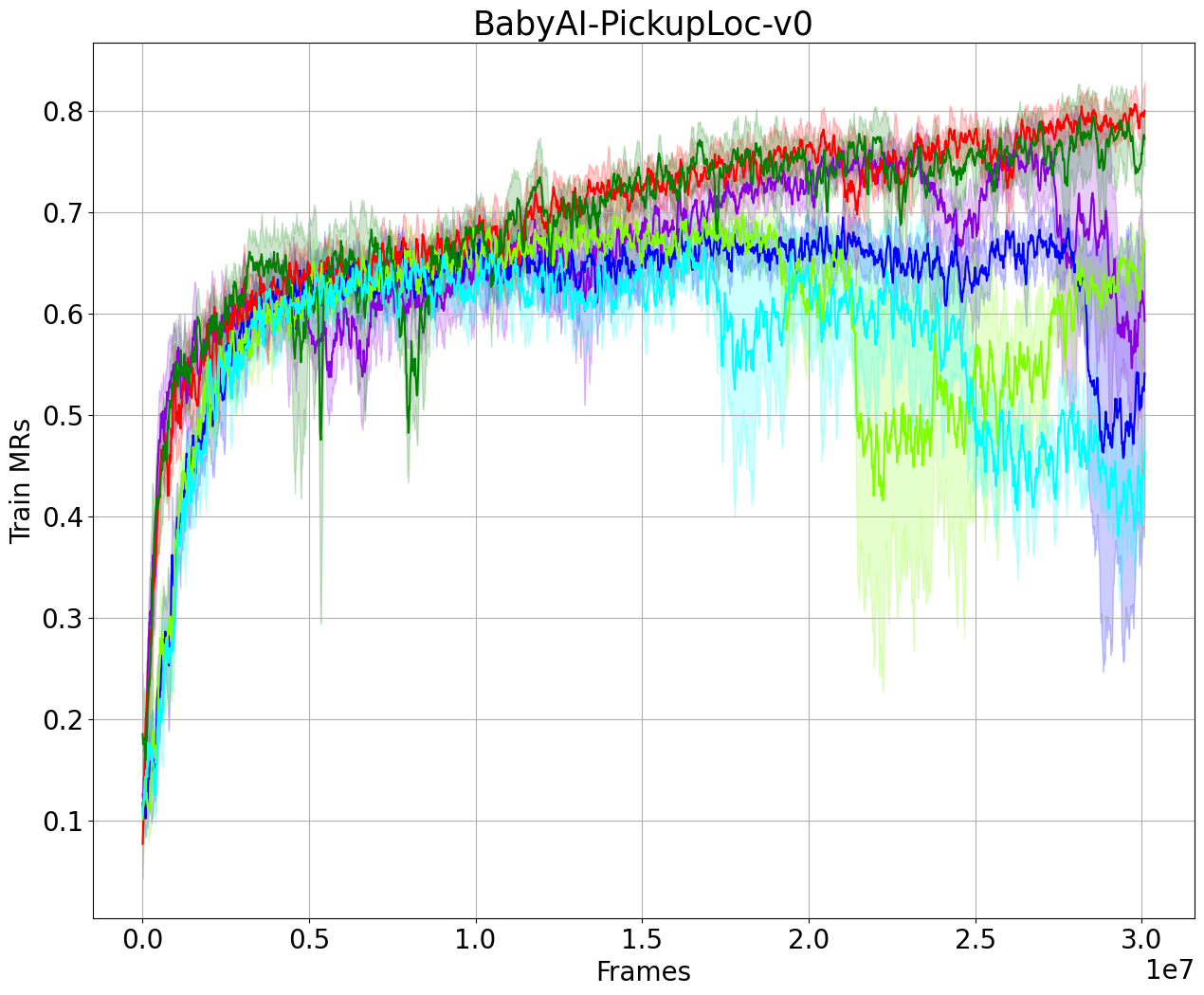}
 \caption{}
 \label{plots-memo-f}
 \end{subfigure}%
 
 \caption{Test and train MR trends comparing memory ablations}
 \label{plots-memo}
\end{figure}

\begin{figure}[!htbp]

 \begin{subfigure}{0.5\textwidth}
 \centering
 \includegraphics[width=\linewidth]{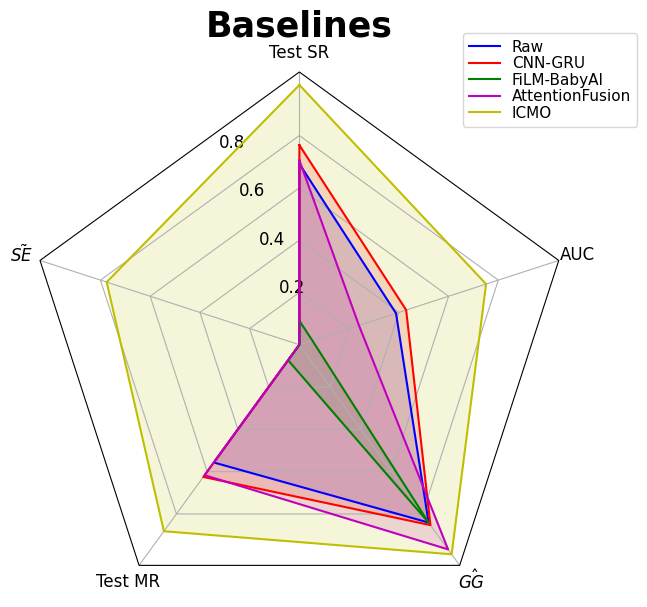}
 \caption{}
 \label{plots-radar-a}
 \end{subfigure}%
 \hfill
 \begin{subfigure}{0.5\textwidth}
 \centering
 \includegraphics[width=0.9\linewidth]{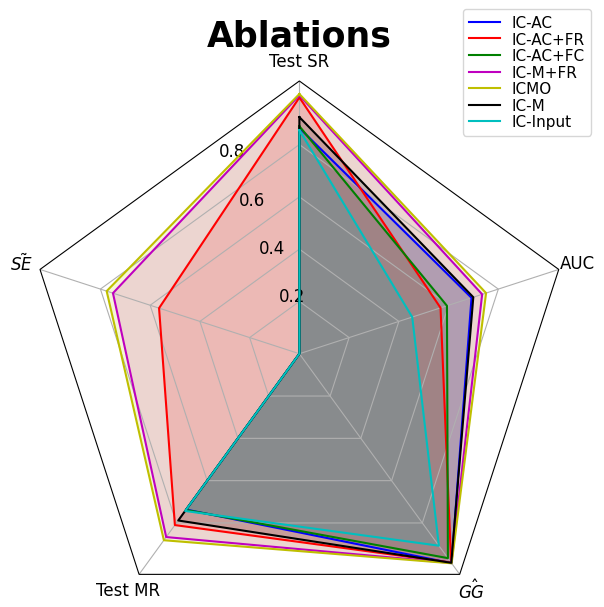}
 \caption{}
 \label{plots-radar-b}
 \end{subfigure}%

 \caption{Test-time Radar Charts indicating the overall performance of (a) baseline models and (b) ablation models against ICMO at a glance}
 \label{plots-radar}
\end{figure}

Table \ref{table-instruction} and Fig. \ref{plots-instr} compare language participation techniques and briefly 
suggests to apply the instruction embeddings in the late stages of the model, like memory or actor-critic networks. Ablations results for memory feedback are represented in Table \ref{table-memory} and Fig. \ref{plots-memo}. We also accumulate the results as radar charts in Fig. \ref{plots-radar} and compare the models in terms of normalized metrics (See \ref{metrics-sec}). From these ablations, we can conclude that the memory feedback to rule or context selection attention with language input to memory (corresponding to IC-M-FR or IC-M-FC, respectively) leads to superior overall performance on the BabyAI levels, supporting our claim on the effect of language-grounded memory and its feedback to lower-level modules discussed in \ref{architecture}.

\begin{table}[h!]
\renewcommand*{\arraystretch}{1.2}
\footnotesize
\centering
\caption{Comparison between the proposed models and the baselines according to test SR, test MR, GG, SE($\alpha=0.9$) over test SRs (divided by $1e6$), and AUC-MR ("$-$" in SE values means the agent didn’t achieve or preserve the desired SR. Best models are emphasized in \textbf{bold} style for each environment)}
\label{table-baselines}
\begin{tabularx}{0.9\textwidth}{>{\centering\arraybackslash}X >{\centering\arraybackslash}X >{\centering\arraybackslash}X >{\centering\arraybackslash}X >{\centering\arraybackslash}X >{\centering\arraybackslash}X}
\toprule
   \multicolumn{6}{c}{\textbf{Test SR}}        \\ \hline 
Env.   & Raw  & CNN-GRU & FiLM-BabyAI & AttentionFusion & ICMO (ours)  \\ \hline
\texttt{ActionObjDoor} & $0.42 \pm 0.02$ & $0.46 \pm 0.10$ & $0.04 \pm 0.01$ & $0.77 \pm 0.05$ & $\mathbf{1.00 \pm 0.00}$ \\
\texttt{GoToSeq} & $0.95 \pm 0.02$ & $0.99 \pm 0.01$ & $0.14 \pm 0.03$ & $0.80 \pm 0.11$ & $\mathbf{1.00 \pm 0.00}$ \\
\texttt{PickupLoc} & $0.70 \pm 0.14$ & $0.83 \pm 0.14$ & $0.09 \pm 0.04$ & $0.55 \pm 0.06$ & $\mathbf{0.98 \pm 0.00}$ \\
\texttt{PutNextLocal} & $0.93\pm0.01$
& $0.89\pm0.09$
& $0.16\pm0.16$
& $0.95\pm0.03$
& $\mathbf{0.99\pm0.01}$ \\
\texttt{OpenDoorsOrder} & $\mathbf{1.00\pm0.00}$
& $\mathbf{1.00\pm0.00}$
& $0.01\pm0.01$
& $0.98\pm0.02$
& $\mathbf{1.00\pm0.00}$ \\
\texttt{Synth} & $0.51\pm0.02$
& $0.28\pm0.01$
& $0.13\pm0.03$
& $0.54\pm0.19$
& $\mathbf{0.86\pm0.04}$
\\\hline
\textbf{Average} &$0.75\pm0.03$
& $0.74\pm0.06$
& $0.09\pm0.05$
& $0.76\pm0.08$
& $\mathbf{0.97\pm0.01}$ \\ \toprule
   \multicolumn{6}{c}{\textbf{Test MR}}        \\ \hline
\texttt{ActionObjDoor} & $0.40 \pm 0.04$ & $0.43 \pm 0.09$ & $0.04 \pm 0.01$ & $0.74 \pm 0.05$ & $\mathbf{0.97 \pm 0.00}$ \\
\texttt{GoToSeq} & $0.82 \pm 0.01$ & $0.86 \pm 0.00$ & $0.12 \pm 0.02$ & $0.67 \pm 0.10$ & $\mathbf{0.88 \pm 0.00}$ \\
\texttt{PickupLoc} & $0.46 \pm 0.10$ & $0.58 \pm 0.09$ & $0.05 \pm 0.03$ & $0.43 \pm 0.05$ & $\mathbf{0.80 \pm 0.02}$ \\ 
\texttt{PutNextLocal} & $0.53\pm0.00$
& $0.61\pm0.10$
& $0.12\pm0.12$
& $0.74\pm0.06$
& $\mathbf{0.82\pm0.01}$
\\
\texttt{OpenDoorsOrder} & $0.95\pm0.01$
& $0.95\pm0.01$
& $0.00\pm0.00$
& $0.95\pm0.01$
& $\mathbf{0.96\pm0.00}$
\\
\texttt{synth} & $0.44\pm0.02$
& $0.26\pm0.02$
& $0.12\pm0.03$
& $0.46\pm0.15$
& $\mathbf{0.69\pm0.03}$
\\\hline
\textbf{Average} & $0.60\pm0.03$
& $0.62\pm0.05$
& $0.08\pm0.04$
& $0.67\pm0.07$
&$\mathbf{0.85\pm0.01}$ \\ \toprule
 \multicolumn{6}{c}{\textbf{GG}}   \\ \hline 
\texttt{ActionObjDoor} & $0.45 \pm 0.01$ & $0.45 \pm 0.10$ & $0.70 \pm 0.01$ & $0.21 \pm 0.05$ & $\mathbf{-0.01 \pm 0.01}$ \\
\texttt{GoToSeq} & $0.06 \pm 0.01$ & $0.02 \pm 0.00$ & $0.42 \pm 0.11$ & $0.16 \pm 0.04$ & $\mathbf{0.01 \pm 0.00}$ \\
\texttt{PickupLoc} & $0.11 \pm 0.06$ & $0.03 \pm 0.01$ & $0.42 \pm 0.03$ & $0.14 \pm 0.02$ & $\mathbf{-0.01 \pm 0.02}$ \\ \texttt{PutNextLocal}
& $0.04\pm0.03$
& $\mathbf{-0.01\pm0.01}$
& $0.04\pm0.02$
& $0.00\pm0.01$
& $0.00\pm0.01$ \\
\texttt{OpenDoorsOrder} & $\mathbf{0.00\pm0.00}$
& $\mathbf{0.00\pm0.00}$
& $0.61\pm0.17$
& $0.02\pm0.01$
& $\mathbf{0.00\pm0.00}$
\\
\texttt{Synth} & $0.19\pm0.05$
& $0.20\pm0.06$
& $0.14\pm0.03$
& $0.06\pm0.05$
& $\mathbf{0.04\pm0.03}$
\\\hline
\textbf{Average} & $0.14\pm0.03$
& $0.12\pm0.03$
& $0.39\pm0.06$
& $0.10\pm0.03$
& $\mathbf{0.01\pm0.01}$
\\ \toprule
 \multicolumn{6}{c}{\textbf{SE Test SR ($\alpha=0.9$)}}      \\ \hline
\texttt{ActionObjDoor} & $-$  & $-$  & $-$  & $-$  & $\mathbf{0.97 \pm 0.32}$ \\
\texttt{GoToSeq} & $17.61 \pm 0.32$ & $10.89 \pm 1.92$ & $-$ & $-$ & $\mathbf{3.53 \pm 0.32}$ \\
\texttt{PickupLoc} & $-$  & $-$  & $-$  & $-$  & $\mathbf{22.73 \pm 2.24}$ \\ \texttt{PutNextLocal} & $10.89\pm1.92$
& $-$
& $-$
& $12.49\pm0.96$
& $\mathbf{4.17\pm0.96}$
 \\
\texttt{OpenDoorsOrder} & $3.53\pm0.32$
& $1.29\pm0.00$
& $-$
& $1.61\pm0.32$
& $\mathbf{0.01\pm0.00}$
\\
\texttt{Synth} & $-$
& $-$
& $-$
& $-$
& $-$
\\\hline
\textbf{Average$^a$} & $-$
& $-$
& $-$
& $-$
& $\mathbf{6.28 \pm 0.77}$ 
\\ \toprule
 \multicolumn{6}{c}{\textbf{AUC-MR}}      \\ \hline
\texttt{ActionObjDoor} & $0.17\pm0.00$ & $0.22\pm0.02$ & $0.04\pm0.01$ & $0.18\pm0.00$ & $\mathbf{0.81\pm0.02}$ \\
\texttt{GoToSeq} & $0.48\pm0.02$ & $0.55\pm0.0.2$ & $0.04\pm0.01$ & $0.33\pm0.02$ & $\mathbf{0.78\pm0.01}$ \\
\texttt{PickupLoc} & $0.52\pm0.01$ & $0.51\pm0.06$ & $0.08\pm0.01$ & $0.21\pm0.01$ & $\mathbf{0.67\pm0.00}$ \\ 
\texttt{PutNextLocal} & $0.41\pm0.00$ & $0.31\pm0.08$ & $0.01\pm0.01$ & $0.51\pm0.02$ & $\mathbf{0.66\pm0.01}$ \\
\texttt{OpenDoorsOrder} & $0.49\pm0.00$ & $0.75\pm0.00$ & $0.00\pm0.00$ & $0.70\pm0.01$ & $\mathbf{0.86\pm0.01}$\\
\texttt{Synth} & $0.24\pm0.01$ & $0.15\pm0.01$ & $0.06\pm0.01$ & $0.22\pm0.05$ & $\mathbf{0.53\pm0.01}$ \\\hline
\textbf{Average} & $0.39\pm0.01$
& $0.42\pm0.03$
& $0.04\pm0.01$
& $0.36\pm0.02$
& $\mathbf{0.72\pm0.01}$
\\ \bottomrule
\end{tabularx}
\footnotesize{$^a$ The \texttt{Synth} environment is leftout in this averaging because it does not converge to a test SR of 0.9.}\\
\end{table}

\begin{table}[h!]
\renewcommand*{\arraystretch}{1.2}
\footnotesize
\caption{\textcolor{black}{Comparison on ablation results of the instruction entrance in the ICMO according to test SR, test MR, GG, SE($\alpha=0.9$) over test SRs (divided by $1e6$), and AUC-MR ("$-$" in SE values means the agent didn’t achieve or preserve the desired SR. Best models are emphasized in \textbf{bold} style for each environment)}}
\label{table-instruction}
\begin{tabularx}{0.8\textwidth}{>{\centering\arraybackslash}X >{\centering\arraybackslash}X >{\centering\arraybackslash}X >{\centering\arraybackslash}X}
\toprule
   \multicolumn{4}{c}{\textbf{Test SR}}        \\ \hline 
Env.   & IC-AC  & IC-M  & IC-Input \\ \hline
\texttt{ActionObjDoor} & $\mathbf{0.99 \pm 0.01}$ & $\mathbf{0.99 \pm 0.01}$ & $0.89 \pm 0.05$ \\
\texttt{GoToSeq} & $\mathbf{1.00 \pm 0.00}$ & $0.98 \pm 0.01$ & $0.80 \pm 0.09$ \\
\texttt{PickupLoc} & $0.59 \pm 0.01$ & $0.74 \pm 0.05$ & $\mathbf{0.88 \pm 0.01}$ \\ \hline
\textbf{Average} 
& $0.86\pm0.00$
& $\mathbf{0.90\pm0.02}$
& $0.85\pm0.05$
\\ \toprule
   \multicolumn{4}{c}{\textbf{Test MR}}        \\ \hline
\texttt{ActionObjDoor} & $0.95 \pm 0.01$ & $\mathbf{0.96 \pm 0.01}$ & $0.85 \pm 0.05$ \\
\texttt{GoToSeq} & $\mathbf{0.88 \pm 0.00}$ & $0.84 \pm 0.02$ & $0.68 \pm 0.07$ \\
\texttt{PickupLoc} & $0.38 \pm 0.03$ & $0.55 \pm 0.03$ & $\mathbf{0.70 \pm 0.06}$ \\ \hline
\textbf{Average} 
& $0.74\pm0.01$ 
& $\mathbf{0.79\pm0.02}$
&$ 0.74\pm0.06$
\\ \toprule
   \multicolumn{4}{c}{\textbf{GG}}         \\ \hline 
\texttt{ActionObjDoor} & $\mathbf{0.00 \pm 0.00}$ & $\mathbf{0.00 \pm 0.00}$ & $0.11 \pm 0.05$ \\
\texttt{GoToSeq} & $0.01 \pm 0.01$ & $0.01 \pm 0.01$ & $0.12 \pm 0.02$ \\
\texttt{PickupLoc} & $0.03 \pm 0.03$ & $0.03 \pm 0.06$ & $\mathbf{-0.03 \pm 0.06}$ \\ \hline
\textbf{Average}
& $\mathbf{0.01\pm0.01}$
& $0.01\pm0.02$
& $0.07\pm0.05$
\\ \toprule
   \multicolumn{4}{c}{\textbf{SE Test SR ($\alpha=0.9$)}}       \\ \hline
\texttt{ActionObjDoor} & $\mathbf{0.65 \pm 0.00}$ & $0.97 \pm 0.32$ & $-$  \\
\texttt{GoToSeq} & $\mathbf{4.81 \pm 3.52}$ & $6.41 \pm 0.64$ & $-$  \\
\texttt{PickupLoc} & $-$ & $-$ & $-$  \\ \hline
\textbf{Average} 
& $-$
& $-$
& $-$
\\\toprule
   \multicolumn{4}{c}{\textbf{AUC-MR}}       \\ \hline
\texttt{ActionObjDoor} & $\mathbf{0.81 \pm 0.01}$ & $0.80 \pm 0.02$ & $0.50 \pm 0.06$  \\
\texttt{GoToSeq} & $\mathbf{0.75 \pm 0.06}$ & $0.68 \pm 0.01$ & $0.37 \pm 0.15$  \\
\texttt{PickupLoc} & $0.52 \pm 0.03$ & $\mathbf{0.62 \pm 0.02}$ & $0.49 \pm 0.09$  \\ \hline
\textbf{Average} 
& $0.69\pm0.03$
& $\mathbf{0.70 \pm 0.02}$
& $0.45 \pm 0.10$
\\ \bottomrule
\end{tabularx}
\end{table}

\begin{table}[h!]
\renewcommand*{\arraystretch}{1.2}
\footnotesize
\centering
\caption{\textcolor{black}{Comparison on the ablation results of the role of memory feedback in ICMO according to test SR, test MR, GG, SE($\alpha=0.9$) over test SRs (divided by $1e6$), and AUC-MR ("$-$" in SE values means the agent didn’t achieve or preserve the desired SR. Best models are emphasized in \textbf{bold} style for each environment)}}
\label{table-memory}
\begin{tabularx}{\textwidth}{>{\centering\arraybackslash}X >{\centering\arraybackslash}X >{\centering\arraybackslash}X >{\centering\arraybackslash}X >{\centering\arraybackslash}X >{\centering\arraybackslash}X >{\centering\arraybackslash}X}
\toprule
\multicolumn{7}{c}{\textbf{Test SR}} \\ 
\hline
Env.   & IC-AC  & IC-AC+FC  & IC-AC+FR  & IC-M  & IC-M+FC (ICMO)  & IC-M+FR  \\ \hline
\texttt{ActionObjDoor} & $0.99 \pm 0.01$ & $0.98 \pm 0.01$ & $0.98 \pm 0.01$ & $0.99 \pm 0.01$ & $\mathbf{1.00 \pm 0.00}$ & $\mathbf{1.00 \pm 0.00}$ \\
\texttt{GoToSeq} & $\mathbf{1.00 \pm 0.00}$ & $0.99 \pm 0.01$ & $0.99 \pm 0.01$ & $0.98 \pm 0.01$ & $\mathbf{1.00 \pm 0.00}$ & $0.99 \pm 0.00$ \\
\texttt{PickupLoc} & $0.59 \pm 0.01$ & $0.64 \pm 0.33$ & $0.96 \pm 0.02$ & $0.74 \pm 0.05$ & $\mathbf{0.98 \pm 0.00}$ & $0.97 \pm 0.02$ \\ \hline
\textbf{Average} 
& $ 0.86\pm0.00$
& $0.87\pm0.11$
& $0.98\pm0.01$
& $0.90\pm0.02$ 
& $\mathbf{0.99\pm0.00}$
& $0.99\pm0.01$
\\
\toprule
\multicolumn{7}{c}{\textbf{Test MR}} \\ 
\hline
\texttt{ActionObjDoor} & $0.95 \pm 0.01$ & $0.93 \pm 0.01$ & $0.93 \pm 0.01$ & $0.96 \pm 0.01$ & $\mathbf{0.97 \pm 0.00}$ & $\mathbf{0.97 \pm 0.00}$ \\ 
\texttt{GoToSeq} & $\mathbf{0.88 \pm 0.00}$ & $0.87 \pm 0.01$ & $0.86 \pm 0.01$ & $0.84 \pm 0.02$ & $\mathbf{0.88 \pm 0.00}$ & $0.86 \pm 0.00$ \\ 
\texttt{PickupLoc} & $0.38 \pm 0.03$ & $0.42 \pm 0.22$ & $0.64 \pm 0.01$ & $0.55 \pm 0.03$ & $\mathbf{0.80 \pm 0.02}$ & $0.77 \pm 0.03$ \\ \hline
\textbf{Average} 
& $0.74\pm0.01$
& $0.74\pm0.08$
& $0.81\pm0.01$
& $0.79\pm0.02$
& $\mathbf{0.88\pm0.01}$
& $0.87\pm0.01$
\\
\toprule
\multicolumn{7}{c}{\textbf{GG}} \\ 
\hline 
\texttt{ActionObjDoor} & $0.00 \pm 0.00$ & $0.03 \pm 0.01$ & $0.01 \pm 0.01$ & $0.00 \pm 0.00$ & $\mathbf{-0.01 \pm 0.01}$ & $0.00 \pm 0.00$ \\ 
\texttt{GoToSeq} & $0.01 \pm 0.01$ & $0.03 \pm 0.02$ & $0.02 \pm 0.00$ & $0.01 \pm 0.01$ & $\mathbf{0.01 \pm 0.00}$ & $0.02 \pm 0.02$ \\ 
\texttt{PickupLoc} & $0.03 \pm 0.03$ & $0.12 \pm 0.09$ & $0.05 \pm 0.03$ & $0.03 \pm 0.06$ & $\mathbf{-0.01 \pm 0.02}$ & $0.02 \pm 0.02$ \\ \hline 
\textbf{Average}
& $0.01\pm0.01$
& $0.06\pm0.04$
& $0.03\pm0.02$
& $0.01\pm0.02$
& $\mathbf{0.00\pm0.01}$
& $0.01\pm0.01$
\\
\toprule
\multicolumn{7}{c}{\textbf{SE Test SR ($\alpha=0.9$)}} \\ 
\hline 
\texttt{ActionObjDoor} & $\mathbf{0.65 \pm 0.00}$ & $4.17 \pm 0.32$ & $4.81 \pm 0.32$ & $0.97 \pm 0.32$ & $0.97 \pm 0.32$ & $0.97 \pm 0.32$ \\
\texttt{GoToSeq} & $4.81 \pm 3.52$ & $8.01 \pm 0.32$ & $10.25 \pm 1.28$ & $6.41 \pm 0.64$ & $\mathbf{3.53 \pm 0.32}$ & $7.05 \pm 0.64$ \\ 
\texttt{PickupLoc} & $-$ & $-$ & $27.53 \pm 0.64$ & $-$ & $22.73 \pm 2.24$ & $\mathbf{21.13\pm5.76}$ \\ \hline
\textbf{Average} 
& $-$
& $-$
& $14.20\pm0.75$
& $-$
& $\mathbf{9.08\pm0.96}$
& $9.72\pm2.24$
\\
\toprule
\multicolumn{7}{c}{\textbf{AUC-MR}} \\ 
\hline 
\texttt{ActionObjDoor} & $\mathbf{0.81 \pm 0.01}$ & $0.54 \pm 0.01$ & $0.48 \pm 0.01$ & $0.80 \pm 0.02$ & $0.81 \pm 0.02$ & $0.79 \pm 0.01$ \\
\texttt{GoToSeq} & $0.75 \pm 0.06$ & $0.65 \pm 0.00$ & $0.68 \pm 0.01$ & $0.68 \pm 0.01$ & $\mathbf{0.78 \pm 0.01}$ & $0.76 \pm 0.01$ \\ 
\texttt{PickupLoc} & $0.52 \pm 0.03$ & $0.58 \pm 0.02$ & $0.54 \pm 0.03$ & $0.62 \pm 0.02$ & $\mathbf{0.67 \pm 0.00}$ & $0.66 \pm 0.03$ \\ \hline
\textbf{Average} 
& $0.69\pm0.03$
& $0.59\pm0.01$
& $0.57\pm0.02$
& $0.70\pm0.02$
& $\mathbf{0.75\pm0.01}$
& $0.74\pm0.02$
\\
\bottomrule
\end{tabularx}
\end{table}

\section{Discussion}
\label{sec:disc}
Regarding Fig. \ref{plots}, the performance gap between ICMO and the baselines is significant. However, previously most involved language-observation fusion structure, i.e. FiLM \citep{madan_fast_2021, brohan2022rt} indicates very poor performance especially on the test split. Comparison to Raw model indicates consistent superiority of ICMO which might arise from meaningful processings carried out by the model. These processings ground the language in memory due to IC-M part, leading to representations that accumulate the history of agent's observations combined with the language description of its goal. In terms of GG in Table \ref{table-baselines} which directly describes the compostional generalization capability of the models, ICMO is the only model that shows near-zero gap whereas in the other models, this gap is meaningful. Moreover, the proposed model manages to reach a test SR of 0.9 and preserve it during training in most environments, while the baselines fail to do so. 

In terms of language participation, from Table \ref{table-instruction} and Fig. \ref{plots-instr}, we can conclude that IC-M and IC-AC are superior compared to IC-Input which can indicate that language involvement in later layers of the model is more desired and the observations need to be processed before alignment with language. \textcolor{black}{Also, by observing the learning curves in Fig. \ref{plots} and Fig. \ref{plots-instr}, we can conclude that early language fusion (as in FiLM-BabyAI and IC-Input) worsens the generalization gap, confirmed by Tables \ref{table-baselines} and \ref{table-instruction}.}
When combined with memory feedback (See Table \ref{table-memory} and Fig. \ref{plots-memo}), passing the instruction embeddings to actor-critic networks instead of the memory, deteriorates the performance of the model, suggesting an effective role for the language in shaping the agent's memory such that it can be used in mid-level processings which determine the activation of inner modules, i.e. rules, or the participation of inner representations e.g. contextual slots in a selective way.

The memory ablations reported in Fig. \ref{plots-memo} and Table \ref{table-memory} confirm that 1) involvement of language as an input to the memory is helpful, and 2) adding memory feedback boosts the agent's performance as well as its training stability (Compare ICMO and IC-M-FR to IC-M in plots \ref{plots-memo-f} and \ref{plots-memo-c}). Feedback to rule selection (IC-M-FR) and to contextual slot selection (IC-M-FC) indicate close performances, but the latter seems to be slightly more successful and indicates less variance. So, the final model that we propose in this paper as ICMO, is IC-M-FC. However, the important aspect is the involvement of language in memory and heading down its feedback to mid-level processings of the model.


\section{Related Work}
\label{apprelatedworks}

\subsection{Language-informed Studies} There have been various language-informed studies in the sequential decision-making setting \citep{luketina_survey_2019, geffner_target_2022, roder_embodied_2021}. \citep{luketina_survey_2019} have provided a survey on language-informed studies in RL, categorizing them into language-conditioned methods, where the language is a part of the main problem formulation and its involvement is mandatory \citep{cote_textworld_2018} like instruction following settings \citep{fu_language_2018, wang_grounding_2021, bahdanau_learning_2018, mirchandani2021ella, madan_fast_2021} and language-assisted methods where the task can be solved without language information, but it can be solved easier using linguistic information \citep{jiang_language_2019, zhong_rtfm_2020, goyal_using_2019}. The participation of the language modality in sequential decision-making settings has been done either by conditioning the policy on language \citep{wang_grounding_2021, chevalier-boisvert_babyai_2018, zhong_rtfm_2020} or by learning auxiliary rewards from language \citep{mirchandani2021ella, goyal_using_2019}. These approaches have been applied to different sequential decision-making problems, such as Hierarchical RL \citep{jiang_language_2019}, Inverse RL \citep{fu_language_2018}, Multi-task RL \citep{chevalier-boisvert_babyai_2018}, and IL \citep{co-reyes_guiding_2018, hejna_iii_improving_2021, shah_lm-nav_2022}. Also, some studies \citep{roder_embodied_2021} from the cognitive neuroscience side have emphasized the importance of grounding language in other input modalities, e.g., vision and policy. Inspired by language learning in children, \citep{roder_embodied_2021} propose to separate language-grounding from low-level skill acquisition. This approach is exemplified in \citep{akakzia_grounding_2021}, where the authors separate language grounding from policy learning using a contextual representation of goals specified in the instruction. 

There have been also studies on using pre-trained models in goal reaching scenarios \citep{paischer2023semantic}, especially where Large Language Models are leveraged for high-level planning \citep{ahn2022can, huang2022inner}. Their results rather reveal the necessity of proper alignment with the environment and the need for grounding non-linguistic modalities in language to gain better understanding of the agent's state, leading to enhanced overall performance.

\subsection{Out of Distribution Generalization in RL} \citep{malik_when_2021} show that despite our intuition, an overall similarity among test and train environments does not yield generalization to unseen scenarios. They propose provable and structurally sufficient conditions for efficient generalization to unseen environments. However, most of the studies focus on empirical methods \citep{kirk_survey_2023}. \citep{kirk_survey_2023} have surveyed the empirical studies on zero-shot generalization in reinforcement learning. They conclude that most of these methods rely on techniques for out-of-distribution generalization in supervised learning, such as invariant learning \citep{agarwal_contrastive_2021, zhang_invariant_2020, zhang_learning_2021}, data augmentation \citep{kostrikov_image_2021, zhang_generalization_2021}, domain randomization \citep{peng_sim--real_2018, openai_solving_2019}, environment generation \citep{wang_paired_2019}, online adaptation including meta-RL methods \citep{nagabandi_learning_2019, duan_rl2_2016, mishra_simple_2018, zintgraf_varibad_2020}, and regularization methods \citep{cobbe2019quantifying}. In terms of inductive biases of language, \citep{hill_human_2020} have applied large pre-trained language models as a source of prior knowledge about tasks to generalize from auxiliary synthetic sentences to human sentences. In the current study, we also leverage the natural language's prior knowledge and inductive biases to better generalize to unseen tasks. 



\subsection{Neuroscientific Studies} A huge body of research proposes that different sup-populations in the brain demonstrate specialization in specific domains which shows a modularity structure \citep{meunier_modular_2010, perich_rethinking_2020, power_functional_2011, sporns_modular_2016, wang_brain_2016, yang2019task, driscoll2022flexible}. One can consider two ways of expressing inductive biases in the brain: \textit{structural}, relating to the configuration of modules, and \textit{functional}, meaning the ability to perform certain aspects of a task \citep{marton2021efficient}. Specifically for visual processing, according to the theory of visual modularity, many qualities of visual perception (such as shape, color, texture, motion, etc.) result from independent processes that take place in diverse cortical and subcortical areas of the brain \citep{calabretta200514evolutionary}. In addition, it is well-documented that the brain’s modules operate and communicate in a sparse regime, giving rise to flexibility in human perception and cognition \citep{jaaskelainen_sparse_2022}. This structural and functional modularity may lead to compositional generalization in cognitive and behavioral levels. In this regard, a number of studies have investigated generalization in biological agents \citep{ito2022compositional, gonzalez2020attentional, marton2021efficient, franklin2020generalizing, riveland2022generalization}.

As for language understanding and representation in the brain, it is well-known that context coding (commonly expressed in natural language) happens in a part of the PFC, named the Frontoparietal Network (FPN), through a process called proceduralization, a multi-step process in which the FPN first encodes the instructional data into declarative code \citep{muhle-karbe_neural_2017}. Then declarative representations are converted into an efficient representation to do the task once this data becomes behaviorally relevant \citep{gonzalez-garcia_frontoparietal_2021}.

Recently, several studies have tried to reveal computation principles behind strong adaptation and compositional generalization in the brain in the presence of multimodal information in the form of instruction and visual input \citep{ito2022compositional, franklin2020generalizing, riveland2022generalization}. More precisely, \citep{ito2022compositional} introduce an experiment to assess compositional generalization for unseen instruction in a zero-shot regime. They suggest that mixed selectivity of abstract variables in a high dimensional space of neural activity -parallel abstract representation- results in the highly adaptive behavior of participants. \citep{riveland2022generalization} also claims that linguistic information by itself can immediately reconfigure the sensorimotor network by modulating certain pathways, leading to generalization to novel tasks. They proposed PFC as the main region responsible for tuning this process.


Moreover, a number of studies mentioned the important role of PFC in action-oriented and reward-driven tasks. Holistically, PFC act as a high-level information aggregator \citep{wang2018prefrontal}, which is closely connected with hippocampus, associated with episodic and semantic memory \citep{eichenbaum2017prefrontal}, and subcortical regions involved in action-oriented tasks such as striatum \citep{neftci2019reinforcement}. PFC is also known to be involved in essential cognitive functionalities such as learning a model of the environment, forming exploratory behaviors \citep{russin2020deep}, and planning \citep{miller2021multi}.

Interestingly, in addition to natural language understanding and goal-directed tasks, PFC is also the main region associated with WM having an important role in selective attention \citep{nobre2019premembering, radulescu_holistic_2019, muhle2021hierarchy} - selective attention refers to the set of functions that prioritize and select information to guide adaptive behavior \citep{nobre2019premembering}. Specifically, the feedback path from PFC to the occipital lobe modulates the activity of mid-level regions of visual processing like the Middle Temporal area (MT) and area V4 \citep{paneri_top-down_2017}.

In summary, the inductive biases injected in the proposed method are consistent with findings in neuroscience about the significant role of PFC in action-oriented tasks and modularity in structural and functional aspects of the brain. This alignment is explained in Section \ref{architecture}.

\section{Limitations and Broader Impact}
Our study aims to enhance instruction-following RL agents to systematically generalize to unseen tasks by leveraging the compositional nature of language. We propose ICMO, a modular architecture with sparse interactions among the network components and the inputs along with memory feedback to improve language grounding in the agent. As stated in Sections \ref{architecture} and \ref{apprelatedworks}, pieces of evidence from neurocognitive science support these inductive biases as they resemble some functionalities of the brain.

In more realistic domains, successful language grounding allows better human-in-the-loop control and human-robot interaction. Although ICMO was experimented against the symbolic BabyAI environment, it emphasizes modularity, sparse interactions, and the role of memory in designing such agents. So, in realistic scenarios, it could promote the reasoning functionalities of the agent.


Although our method is tested against symbolic inputs, it does not make any assumptions about the input structure and can be modified to handle larger observation spaces. Even if the slots are key to its success, one can obtain such high-level and factorized representations using pre-trained encoders for downstream tasks. Slot-Attention \citep{locatello_object-centric_2020} or DINOSAUR \citep{seitzer2022bridging} could be candidates here. Also, there is a line of studies in the language-informed sequential decision-making literature \citep{mirchandani2021ella, carta2022eager, campero2021learning, zhao2021consciousness, loynd2020working} that focus on symbolic environments. Following this line, we propose inductive biases to improve the related baselines.

Since this paper introduces techniques to improve RL agents on a fundamental level, we don’t expect any negative societal impacts.

\section{Conclusion}
We have introduced ICMO, a modular encoder model with sparsely-connected units and a language-conditioned memory which sends task-relevant feedbacks to the mid-level processing of the observations. We have tested this model in the zero-shot systematic generalization setting. 
We compared our method on several challenging tasks in BabyAI environments with strong baselines. Our model could significantly improve systematic generalization and training stability by involving memory feedback in sparse processing of the observation via modular units, and conditioning the memory on language. Besides the inductive biases introduced in this study, there are several future directions which can further improve the current results. Using auxiliary loss functions to induce certain restrictions in the model could be helpful. Employing information bottlenecks in the form of regularization potentially can be effective in generalization. 
Moreover, one can try scenarios with a richer language modality (e.g. descriptive sentences, wikis, etc.) using ICMO and involve different texts (instructive, descriptive, guidance, etc.) using the proposed techniques to maximize information utilization in the agent.

\section{Declarations}

\begin{itemize}
\item \textbf{Funding}: The authors did not receive support from any organization for the submitted work.

\item \textbf{Conflicts of interest/Competing interests}: The authors have no competing interests to declare that are relevant to the content of this article.

\item \textbf{Ethics approval}:Not applicable.

\item \textbf{Consent to participate}: Not applicable.

\item \textbf{Consent for publication}: Not applicable.

\item \textbf{Availability of data and material}: We do not analyse or generate any datasets, because our work proceeds within a fundamental approach examined on publicly available benchmarks \citep{chevalier-boisvert_babyai_2018}.

\item \textbf{Code availability}: Our code is publicly available at \href{https://github.com/nhashemi202/ICMO.git}{https://github.com/nhashemi202/ICMO.git}.

\item \textbf{Authors' contributions}: All authors contributed to the conception of the work. N.H.D., R.R.R., and M.S.B. were involved in designing the study and analyzing the results. N.H.D. played a primary role in designing the experiments, implementing the codes, and analyzing the results. N.H.D. drafted the manuscript, and all authors critically reviewed and revised it.

\end{itemize}

\bibliography{sn-article}

\end{document}